\RequirePackage{fix-cm}
\documentclass{svjour3}                     
\smartqed  
\usepackage{graphicx}
\usepackage[caption=false]{subfig}
\usepackage[export]{adjustbox}
\usepackage[linesnumbered]{algorithm2e}
\usepackage{amsmath}
\usepackage{booktabs}
\usepackage{gb4e} 
\noautomath
\usepackage{etoolbox}
\usepackage{hyperref}

\usepackage{xargs}                      
\usepackage[usenames,dvipsnames,table]{xcolor}
\usepackage[colorinlistoftodos,prependcaption,textsize=tiny,disable]{todonotes}		
\newcommandx{\aUnsure}[2][1=]{\todo[linecolor=red,backgroundcolor=red!25,bordercolor=red,#1]{A--- #2}}
\newcommandx{\aChange}[2][1=]{\todo[linecolor=blue,backgroundcolor=blue!25,bordercolor=blue,#1]{A--- #2}}
\newcommandx{\aInfo}[2][1=]{\todo[linecolor=OliveGreen,backgroundcolor=OliveGreen!25,bordercolor=OliveGreen,#1]{A--- #2}}
\newcommandx{\aImprovement}[2][1=]{\todo[linecolor=Plum,backgroundcolor=Plum!25,bordercolor=Plum,#1]{A--- #2}}

\usepackage{glossaries}
\usepackage{glossary-longbooktabs}

\makeglossaries
\DeclareMathOperator*{\argmin}{arg\,min}

\newcommand{\transAr}[1]{\emph{#1}}
\newcommand{\transArT}[2]{\transAr{#1} `#2'} 
\usepackage{listings}
\begin{document}


\newglossaryentry{abbasid}{name=Abbasid Empire, description={The third of the Islamic caliphates to succeed the Prophet Muhammad. It lasted from 750 to 1258 CE.}}

\newglossaryentry{ah}{name=AH, description={Anno Hegirae. Also known as Islamic Hijri, it is a lunar calendar. 1439AH corresponds to the period from September 2017 to September 2018 AD}}

\newglossaryentry{boilerplate}{name=boilerplate, plural=boilerplates, description={A text fragment that is used in multiple contexts with minimal changes from the original}}
\newglossaryentry{ce}{name=CE, description={Common Era (not sure we need this one)}}
\newglossaryentry{coha}{name=COHA, description={Corpus of Historical American English}}

\newglossaryentry{ca}{name=CA, description={Classical Arabic; the language used in pre-modern texts}}
\newglossaryentry{ce}{name=CE, description={Common Era (not sure we need this one)}}
\newglossaryentry{coha}{name=COHA, description={Corpus of Historical American English}}

\newglossaryentry{exegesis}{name=exegesis, description={A critical explanation of a religious text}}

\newglossaryentry{JK}{name={JK}, description = {Al-Jami' Al-Kabir, a digital library of Arabic texts}}

\newglossaryentry{matreslectionis}{name=\textit{matres lectionis}, description={Consonants used in some Semitic languages to indicate a vowel}}
\newglossaryentry{msa}{name=MSA, description={Modern Standard Arabic}} 

\newglossaryentry{openiti}{name=OpenITI, description={Open Islamicate Texts Initiative}}
\newglossaryentry{openiti core}{name={OpenITI core}, description={OpenITI corpus with one file per work}}
\newglossaryentry{openiti complete}{name={OpenITI full}, description={OpenITI corpus with all files including redundant files}}
\newglossaryentry{pca}{name=PCA, description={Post-Classical Arabic}} 
\newglossaryentry{psca}{name=PSCA, description={Pre-Standardized Classical Arabic}}
\newglossaryentry{sa}{name=SA, description={Standard Arabic}}
\newglossaryentry{sca}{name=SCA, description={Standardized Classical Arabic}}
\newglossaryentry{shamela}{name=Shamela, description ={Al-Maktaba Al-Shamela (``The Complete Library''), a digital library of Arabic texts}}
\newglossaryentry{syriac}{name=Syriac, description={A dialect of Middle Aramaic that appeared in the early 1st century CE}}

\newglossaryentry{svd}{name=SVD, description={Singular value decomposition}}
\newglossaryentry{vnc}{name=VNC, description={Variability-based Neighbor Clustering}}
\newglossaryentry{wenc}{name=WENC, description={Word-Embedding-based Neighbor Clustering}}
\newglossaryentry{hadith}{name=Hadith, description={Reports about sayings and deeds of the Prophet Muhammad and his companions}}
\newglossaryentry{diwan}{name=diwan, plural={diwans}, description={A collection of poetry}}
\newglossaryentry{loess}{name=LOESS, description={A non-parametric regression method.}}

\title{Studying the History of the Arabic Language
\thanks{* = Equal contribution}
}
%

\subtitle{Language Technology and a Large-Scale Historical Corpus}


\author{Yonatan Belinkov*         \and
        Alexander Magidow*        \and 
        Alberto Barr\'on-Cede\~no     \and 
        Avi Shmidman              \and 
        Maxim Romanov             
}

\authorrunning{Belinkov, Magidow, et al.} 

\institute{Y. Belinkov \at
              MIT Computer Science and Artificial Intelligence Laboratory, Cambridge, MA 02139, USA \\
              \email{belinkov@mit.edu}           
           \and
           A. Magidow \at
              Department of Modern and Classical Languages and Literatures, University of Rhode Island, USA \\
              \email{amagidow@uri.edu}
            \and 
              A. Barr\'on-Cede\~no \at 
              Qatar Computing Research Institute, HBKU, Doha, Qatar \\
              \email{albarron@ \{hbku.edu.qa $|$ gmail.com\}}
            \and 
              A. Shmidman \at 
              Department of Hebrew Literature, Bar-Ilan University, Israel \\
              Dicta: The Israel Center for Text Analysis \\
              \email{shmidman@gmail.com}
            \and 
              M. Romanov \at 
              Department of History, University of Vienna, Vienna, Austria \\
              \email{maxim.romanov@univie.ac.at}
}

\date{Received: date / Accepted: date}

\maketitle

\begin{abstract}

Arabic is a widely-spoken language with a long and rich history, but existing corpora and language technology focus mostly on modern Arabic and its varieties. Therefore, studying the history of the language has so far been mostly limited to manual analyses on a small scale.  In this work, we present a large-scale historical corpus of the written Arabic language, spanning 1400 
years. We describe our efforts to clean and process this corpus using Arabic NLP tools, including the identification of reused text. We study the history of the Arabic language using a novel automatic periodization algorithm, as well as other techniques. Our findings confirm the established division of written Arabic into Modern Standard and Classical Arabic, and confirm other established periodizations, while suggesting that written Arabic may be divisible into still further periods of development.

\keywords{Arabic \and Corpus \and Periodization \and Text Reuse \and Historical Linguistics}
\end{abstract}

\section{Introduction}

Language is complex and dynamic. It changes across space and time, from generation to generation. New words are introduced and old words go out of use; words acquire new meanings and change their old meanings; and grammatical norms that existed in the past may become obsolete in the future. 
Language use is partly documented by texts which preserve traces of that change including variations in spelling, prefixes or suffixes that appear or disappear across eras and changes in word meanings.

The Arabic language is no exception to this, but provides a challenge for the historical linguist. Unlike many languages where the standardization of writing was a long and contended process, leaving a trace of that process in the written record, Arabic writing was standardized quite early. This created a divergence between the spoken and written languages and while spoken Arabic has continued to evolve, written Arabic was essentially \textit{fossilized} in the 8th century if not earlier. Between that period and today, spellings vary little, the grammar of formal written Arabic shows few changes from the earliest texts, and aside from a period of modernization in the 19th century, there is little apparent change in the lexicon.

While many studies have been conducted on non-standard texts that do diverge from this norm, or on the history of spoken Arabic, comparatively little attention has been paid to standard written Arabic. In this paper we seek to develop tools that enable us to determine whether formal written Arabic is as unchanging and homogeneous as it first appears, or whether we can divide it into separate periods, analogous to those in other languages (e.g., ``Old English'',``Early Modern English''). With 1400 years of written Arabic texts across many different genres and with very subtle differences between eras, this is a task that is incredibly difficult to undertake using traditional philological methods, which may explain the small number of previous studies. For that reason, we seek to develop computational resources and methods to investigate the periodization of standard written Arabic. Specifically, we seek to answer the following:

\begin{enumerate}  
\item What computational tools and resources are needed to investigate the periodization of Arabic?
\item Can formal written Arabic be divided into temporally-distinct periods based on linguistic evidence?
  \item Are previously proposed periodizations of written Arabic accurate?
\end{enumerate}

After a brief overview of Arabic, and a review of previous studies in this area (Section~\ref{sec:related-work}), 
the rest of the paper is divided in two parts.
In the first part (Section~\ref{sec:corpus}), we describe our efforts to 
process \gls{openiti} --- a large-scale diachronic corpus of Arabic with approximately $1.5~G$ words. \gls{openiti} is the largest publicly-available historical corpus of Arabic that we are aware of. We make the processed corpus available for the community in order to facilitate data-driven research of the history of the language.

In the second part of the paper, we develop computational methods for investigating the history of the Arabic language. In Section~\ref{sec:reuse} we adapt a text reuse algorithm that was previously applied to a relatively small Hebrew/Aramaic corpus in order to identify exact and approximate matches in the large \gls{openiti} corpus. Identifying matches is especially important because many of the documents in the corpus quote and paraphrase  large quantities of texts from earlier works, sometimes many centuries earlier. 
Texts that contain language from very different eras make all forms of historical linguistic analysis more difficult. 

In Section~\ref{sec:periodization} we develop a novel data-driven periodization algorithm that is based on word embeddings. Contrary to previous methods, our algorithm captures language use on the level of full corpora or subsets of corpora, rather than being limited to a handful of linguistic features. We apply this algorithm to the OpenITI historical corpus and find well-known as well as new periodizations. 

In Section~\ref{sec:expert} we utilize the corpus for an expert study of Arabic periodization. First, we verify that written Arabic does indeed change remarkably slowly by tracking the lifespan of Arabic words in contrast to English words. 
We also study several important linguistic phenomena that have so far only been anecdotally analyzed in the literature.

\medskip
The main contributions of our work are:
\begin{itemize}
\item Preprocessing \gls{openiti}, a diverse large-scale historical corpus of Arabic, including morphological segmentation, part-of-speech tags, lemmatization, and syntactic parse trees. 
\item A complete identification of parallel matches in the corpus based on a novel adaptation of a text reuse algorithm. The algorithm is adapted for Arabic and runs efficiently on the large-scale \gls{openiti} corpus. We are able to identify and remove $292~M$ words of reused text, nearly $20\%$ of the total corpus.  
\item A novel periodization algorithm relying on word embeddings, which can be applied to any large-scale historical corpus. We demonstrate its applicability to the Arabic case on \gls{openiti}. 
\item New insights regarding the history of the Arabic language, that illuminate its development from early times to the modern days. In particular, our computational methods affirm the established periodization of Standard Arabic into  Classical and Modern Standard Arabic, and point to new periodizations for Classical Arabic.
\end{itemize}

\section{Linguistic Background} \label{sec:background}

Arabic is a \textit{diglossic} language, meaning there is a single formal language, Standard Arabic (\gls{sa}),\footnote{A glossary with some definitions and further background appears at the end of this manuscript.} which is used as a language of writing and formal communication. Everyday life is conducted in a divergent set of spoken languages, referred to collectively as ``colloquial Arabic''. Written Arabic in various forms is attested for some time prior to Islam, but the coming of Islam marks the beginning of a vast written tradition. Even prior to Islam, \gls{sa} was a relatively homogeneous register used for oral literature \cite{EALLHistArabic}. However the coming of Islam, dated to 622 CE, when the early Muslim community moved from Mecca to the city of Medina, brought a huge increase in written production. \gls{sa} was largely standardized even before the 8th century CE, but that is the time period most strongly associated with the establishment of explicit linguistic standards. Oral texts from prior to that time were committed to writing, but were almost certainly edited later to conform more closely to the standard \cite{EALLMiddleArabic}. Such early standardization means that the \gls{sa} of the 8th century CE is still basically accessible to a reader today --- for example, the collection of stories \emph{Kalila wa Dimna} from ca. 750 CE is considered appropriate reading material at the middle school level today, with archaic terms or structures elucidated by footnotes. Though the literary style of \gls{sa} varies between genres and eras, the basic orthography, morphology, syntax, and even vocabulary appear to have remained largely the same since that time. Impressionistically, there is very little variation between a modern formal  text in Arabic and a text from nearly a thousand years ago, if they cover similar topics. 

This is not to say that all Arabic writing is homogeneous. There are variants of written Arabic which diverge significantly from \gls{sa}, even in the pre-modern era: the language of the Ancient North Arabian inscriptions predates the standardization of Arabic, and probably the coming of Islam, and so differs in alphabet, spelling, and lexicon~\cite{AlJallad15Safaitic}. ``Middle Arabic'', a term which is not chronological, refers to writings which do not match the norms of \gls{sa}, whether they are early Islamic era papyri or Judeo--Arabic letters from the 14th century~\cite{MiddleArabicVolume12}. These texts are of great use for historical linguists, but a relatively small quantity have been digitized. In the case of Ancient North Arabian texts, they are far too different in genre and vocabulary to be easily comparable to \gls{sa} texts. By far the largest body of writing is in \gls{sa}, as this is the language of writing and publishing. Even texts which may have been closer to Middle Arabic at one point were likely corrected by later editors to better conform to the \gls{sa} style~\cite{EALLMiddleArabic}. 

Therefore, a major question, is whether the apparently homogeneous \gls{sa} can be divided into eras in the same manner as many European languages, which generally did not experience standardization until late in their literary history. 
The autochthonous linguistic tradition typically treats \gls{sa} as a single, undifferentiated language, while western Arabists divide \gls{sa} into Classical Arabic (\gls{ca}), the language used in pre-modern texts, and Modern Standard Arabic (\gls{msa}). MSA is said to have arisen due to increased contact with the West, starting in the late 18th and early 19th centuries. The largest changes were in lexis, as Western (French, English, German, Italian) terms were adopted or translated, though there may have also been some changes in syntax  under the influence of European languages, or even of the spoken colloquial Arabic dialects \cite{Newman13MSAOxfordHandbook}.

Rarely is \gls{ca} itself divided into other eras, with most variation in \gls{ca} attributed to register or geography, rather than temporal variation~\cite[e.g., pp. 38--41]{Holes04}. A small quantity of research supports dividing \gls{ca} further. There are two slightly different accounts in the literature.  
Fischer suggests a tri-partite division of pre-Standardized \gls{ca}, Standardized \gls{ca}, and Post-\gls{ca} \cite{EALLClassicalArabic}. Pre-Standardized Classical Arabic (\gls{psca}) is pre-Islamic and early Islamic, primarily attested in quotations of older texts, and represents a very limited corpus. Standardized Classical Arabic (\gls{sca}) fully develops by the 8th century CE, less than two centuries after the coming of Islam, and there are only a small number of minor changes that are claimed to separate it from the Pre-Standardized Era. The period of Post-Classical Arabic (\gls{pca}) is dated to the fall of the \gls{abbasid} at the hands of the Mongols (1258 CE), when the power of Arabic-using regimes in Iraq became decentralized, with power shifting elsewhere \cite{Romanov17Biographical}.  
Ali suggests a slightly different timeline in which Arabic undergoes two `renaissances'~\cite{Ali87Vocab}. The first begins under the \gls{abbasid} in Baghdad (750--1258), witnessing a huge growth in text production and translation of texts from \gls{syriac} and Greek. Following the end of the \gls{abbasid} empire, Arabic experiences a period of decline until the 19th century renaissance that produces \gls{msa}. 

These models make slightly different predictions. Fischer's model has \gls{sca} coming early and being distinguished primarily from pre-Islamic Arabic and very early texts. Ali's model predicts significant changes during the Abbasid period, particularly an increase in vocabulary during this time. Fischer recognizes \gls{pca} as a development of the language, whereas Ali sees it as  the end of a renaissance, and hence would not predict rapid change or growth in \gls{pca}. Both models treat the fall of the Abbasids in 1258 as a pivotal event in the history of Arabic, but it should be emphasized that this is a political change and only limited linguistic evidence has been shown to separate the two eras \cite{Magidow16HaHuna}.

Within this work, we find strong support for the existence of \gls{msa} as separate from \gls{ca}, but within \gls{ca} the results are less definitive. There is very limited evidence for linguistic changes that occur shortly before the fall of the Abbasids. Other evidence supports the Abbasid renaissance model, with a break between the language of the earliest texts and those of the 4th century AH onward.

\section{Related Work} \label{sec:related-work}

In this section we give an overview of existing Arabic corpora as well as some existing methods for  text reuse identification, periodization, and dating. 

\subsection{Arabic Corpora}
\label{sub:related-corpora}
Though there has been increasing interest in compiling Arabic corpora in the past decade, very little work has been done on compiling historical corpora reflecting the long history of the Arabic language.  Most of the existing corpora focus on modern written Arabic texts, particularly online news media, 
although there are a growing number of corpora which feature written and to a lesser degree spoken material from Arabic dialects. We mention here several relevant corpora and refer to other surveys for more details~\cite{Al-Sulaiti:2004,Al-Thubaity:2015:ACK:2812480.2812508,shoufan-alameri:2015:WANLP,zaghouani2014critical}.

To date, only a small number of diachronically oriented corpora of Arabic have been produced and made available. The King Saud University Corpus of Classical Arabic (KSUCCA)~\cite{alrabiah2013design}\footnote{\url{http://ksucorpus.ksu.edu.sa}} 
consists of approximately $50.6~M$ words from the first 4 Islamic centuries. It has been morphologically analyzed with the MADA tool~\cite{Habash:2005,Habash:2009}. 
Almost all of the texts are derived from the Shamela website which is a major source of texts for \gls{openiti}.\footnote{\url{http://shamela.ws}} Text metadata is given by century, so more granular buckets are not possible in the current state of this corpus. The Historical Arabic Corpus (HAC)~\cite{Hammo2016} has about $45~M$ words from diverse time periods, with text data given by century, as well as automatic part-of-speech tagging information.  
Other Classical Arabic corpora that are worth mentioning include a $5~M$ word corpus~\cite{elewa2004}, which does not seem to be publicly available, another $2.5~M$ word corpus~\cite{Rashwan:2011:SAD:2209821.2210698},\footnote{\url{http://www.RDI-eg.com/RDI/TrainingData}} and Tashkeela, a $76~M$ word corpus of texts  from  the Shamela website.\footnote{\url{https://sourceforge.net/projects/tashkeela}} These corpora are either  small or lack high-quality temporal metadata.

Finally, a few large corpora are available only via online search interfaces:
KACST Arabic Corpus~\cite{Al-Thubaity:2015:ACK:2812480.2812508} has more than $700~M$ words, including around $16~M$ words from the beginning of the Islamic era.
The Leeds Arabic Internet Corpus\footnote{\url{http://corpus.leeds.ac.uk/internet.html}} and the International Corpus of Arabic\footnote{\url{http://www.bibalex.org/ica/en/About.aspx}} contain $300~M$ and $100~M$ words, respectively, but they include mostly modern texts. The well-known ArabiCorpus\footnote{\url{http://arabicorpus.byu.edu}} has more than $170~M$ words from diverse periods of time, and arTenTen~\cite{Arts2014357} is a $5.8~G$ word Web corpus, with a sub-corpus of $115~M$ words available through Sketch Engine~\cite{Kilgarriff04thesketch}. There is also CLAUDia~\cite{Milichka14},  
 another Shamela-based corpus, but with added genre metadata; however, only a subset appears to be accessible via a Web interface.\footnote{\url{http://arabiccorpus.com/index.htm}} While these corpora are very large and may contain texts from different periods, they are not directly accessible and lack sufficient diachronic information. 

In contrast to previous resources, our corpus has fine-grained time information, it covers most of the history of the written Arabic language, and it is available for developing NLP applications or supporting digital humanities projects (cf. Section~\ref{sec:corpus}). 

\subsection{Identification of Text Reuse}

A popular approach to text reuse addressed the problem in the context of domains such as newspaper texts and law bills~\cite{CloughACL:02,Smith:2014:DML:2740769.2740800,wilkerson:2015}. 
A standard approach to approximate-matching tasks is the use of edit-distance measures such as Levenshtein Distance (e.g.,~\cite{Scherbinin:2009}); however, such an approach is not efficient, particularly given a corpus of this size.
More successful and efficient models rely on either word- or character-level $n$-grams to align similar chunks of text~\cite{Kasprzak:10,Muhr:10}. Documents are broken down into overlapping sequences of relatively short $n$-grams ($\sim 4$ for words; $\sim 16$ for characters) and the resulting hashes are either indexed for search or compared pairwise in order to find collisions. 

Some text reuse detection models have been open-sourced, such as \texttt{passim}.%
\footnote{Available at \url{https://github.com/dasmiq/passim}; previously applied to Arabic texts: \url{http://kitab-project.org/text-reuse-methods/}.} 
A recent international challenge focused on text reuse detection in Arabic~\cite{Bensalem:15}, but the reuse cases were artificially generated in order to allow for an objective evaluation.
The approaches that addressed the task were mostly based on $n$-grams comparison as well, with some Arabic-specific preprocessing. 
An especially interesting study of real text reuse is~\cite{zemanek-milivcka:2014:CLFL}, which detected quotations in the CLAUDia corpus (cf. Section~\ref{sub:related-corpora}) and built a network of documents based on metadata and quoted texts. However, their method focuses on long verbatim quotations, whereas we are interested in approximately-matching parallel passages with possible variations. 
Instead, we follow a recent approach  for finding parallel passages across a large Hebrew/Aramaic corpus~\cite{shmidman2016identification}, adapt it to the Arabic language, and scale it up to handle the large corpus. 

We refer the reader to~\cite{Clough:2009,li:2016,Smith:2014:DML:2740769.2740800} for a broader overview of text reuse detection models and their applications.

\subsection{Periodization, Text Dating, and Language Change}
Most approaches to periodization, qualitative and quantitative, have either assumed standard periodizations proposed by traditional linguistics, or worked with pre-determined temporal bins (e.g. decades or centuries). Some recent works use clustering algorithms to determine more natural periodizations, but these approaches require pre-selected variables for classification. For example, in a study on automatic clustering of English, frequencies of get-passives and verb conjugation suffixes \mbox{\emph{-(e)th}} and \mbox{\emph{-(e)s}} were used as input for the clustering algorithm \cite{GriesHilpert}. In a study on Chinese, the variables were already selected with an awareness of the history of Chinese morphosyntax, and the variables  themselves were encoded into the annotated corpus \cite{Ji}. Although these methods are promising, they require an existing sense of meaningful variables which vary diachronically, whereas we seek to periodize language without subscribing to pre-defined variables.

There is a fairly large body of work on text dating, especially using clues like time expressions, but also various other features~\cite{chambers:2012:ACL2012,dalli-wilks:2006:ARTE,niculae-EtAl:2014:EACL2014-SP,popescu-strapparava:2015:SemEval}. Previous research operated at different granularity levels and algorithmic methods, including pairwise learning-to-rank, multi-class support vector machines, and language models~\cite{niculae-EtAl:2014:EACL2014-SP,popescu-strapparava:2015:SemEval,jong2005temporal}. These methods aim to learn to assign dates to undated documents. Our main motivation in this paper is different: given a corpus of dated texts, we seek to find a division into historical time periods.

Another line of work has applied computational methods for studying diachronic language change. These methods learn the context in which words appear in large corpora, and define semantic change as a change to that context. Earlier attempts used this principle to detect meaning change in 19th century British English ~\cite{Sagi2009}, or in a more modern corpus ~\cite{Wijaya2011}. However, they tested semantic change for a very small set of words. In contrast, recent work has extended the scope of such analysis, and is now able to automatically detect words with significant semantic change for the entire vocabulary, building on word embeddings learned from large raw texts~\cite{Dubossarsky:2015,P16-1141,kim-EtAl:2014:W14-25,Kulkarni:langchange}. We develop a periodization algorithm based on word embeddings to automatically cluster periods of similar language use.  

\section{Corpus Construction} \label{sec:corpus}

In this section we describe the 
\gls{openiti} corpus as well as the preprocessing we carried out on it in order to perform our computational 
linguistic analyses.

\subsection{Text Collection}

The corpus we present in this work is the Arabic portion of the works collected by the Open Islamicate Texts Initiative (\gls{openiti}), an on-going effort to collect texts in Arabic, Persian, and other languages.\footnote{\url{http://iti-corpus.github.io}} The documents are collected from freely available editions of primarily religious and literary texts from different time periods.  The main sources of these texts include Al-Maktaba Al-Shamela,\footnote{``The Complete Library'', \url{http://shamela.ws}} referred to here as \gls{shamela}, the Shia online library,\footnote{\url{http://shiaonlinelibrary.com}} and Al-Jami' Al-Kabir (\gls{JK}).\footnote{``The Great Collection''. Not available online; it was published on an external hard-drive.} 
The documents were converted 
into a unified format and organized into a machine-readable corpus, which is now openly available.%
\footnote{\url{https://github.com/OpenITI}} 
Accompanying metadata information includes standardized author names, text title and author date of death.\footnote{Standardization of author names and text titles is a difficult task for Arabic and was performed using a combination of digital tools (\url{https://github.com/maximromanov/DuplAway}) and manual inspection.} Though \gls{shamela} includes text genre data, it is not very reliable and so this was not retained. We use the author date of death as the document date throughout this work. Author lifespans average around 70 years, so there is a lag surrounding all of the data that needs to be taken into consideration during analysis~\cite[p. 239]{romanovdiss}. All dates in the corpus are based on the Islamic calendar, which begins in 622 and which uses lunar years. Where reasonable, we have converted these dates to Gregorian years, but where they would produce unrounded bins and boundaries, we have retained the Islamic century numbering. 

\begin{table}[t]
\centering
\subfloat{
\begin{tabular}{l|cc}
\toprule
&	Words	& Documents\\
\midrule
\gls{openiti complete}	& $1.5~G$	& $7,144$	\\
\gls{openiti core}	& $725~M$	& $4,322$ 	\\
\bottomrule
\end{tabular}
}\hspace{1cm}
\subfloat{
\begin{tabular}{l|r}
\toprule
Source	& Documents \\
\midrule
Shamela	& $2,375$	\\
JK	& $1,106$	\\
Shia library	& $823$	\\
Others	& $8$	\\
\bottomrule
\end{tabular}
}
\caption{Statistics on the \gls{openiti} corpus showing the number of words and documents in \gls{openiti complete} vs.  \gls{openiti core} (left), and the distribution of sources in \gls{openiti core} (right). 
\label{tab:openiti-versions}}
\end{table}

The current \gls{openiti} corpus actually retains all duplicate texts. We refer to this as \gls{openiti complete}. De-duplication has systematically chosen an instantiated text for each unique work from the source corpora. We refer to this as \gls{openiti core}. 
Table~\ref{tab:openiti-versions} shows some statistics of the two versions.
After de-duplication, 
\gls{openiti core} has 1106 texts (26\%) from JK, 2375 (55\%) from Shamela, and 823 (19\%) from the Shia library; 8 texts are from elsewhere. 
The de-duplication was conducted in a semi-automatic manner.\footnote{More details are available at: \url{https://maximromanov.github.io/OpenITI/}.}

\begin{figure}[h]
\centering
\subfloat[\gls{openiti complete} document count.]{
\centering
\includegraphics[width=0.5\linewidth]{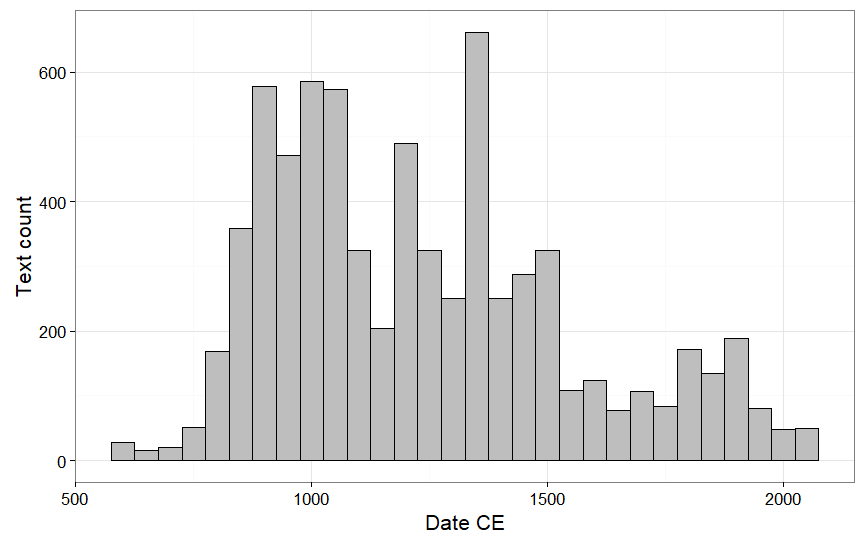}
\label{fig:counts-periods-full-text}
}
\subfloat[\gls{openiti complete} word count.]{
\centering
\includegraphics[width=0.5\linewidth]{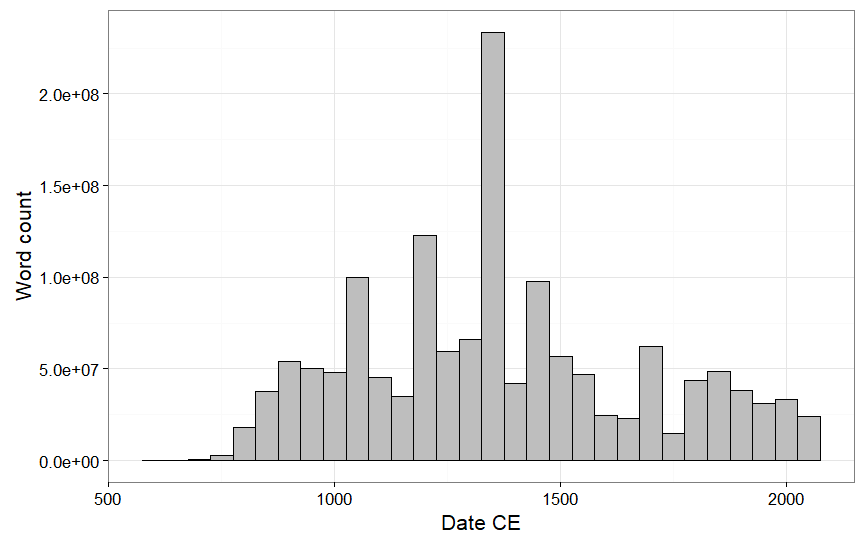}
\label{fig:counts-periods-full-word}
}
\\
\subfloat[\gls{openiti core} document count.]{
\centering
\includegraphics[width=0.5\linewidth]{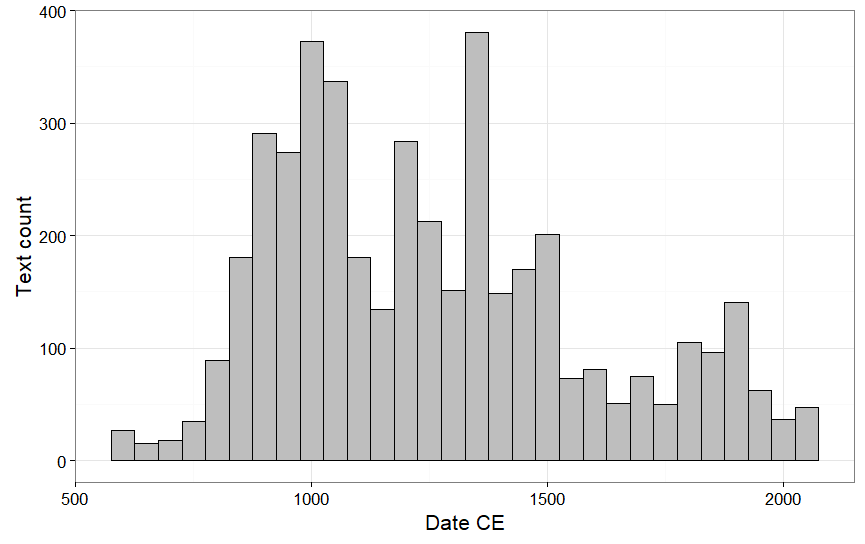}
\label{fig:counts-periods-core-text}
}
\subfloat[\gls{openiti core} word count.]{
\centering
\includegraphics[width=0.5\linewidth]{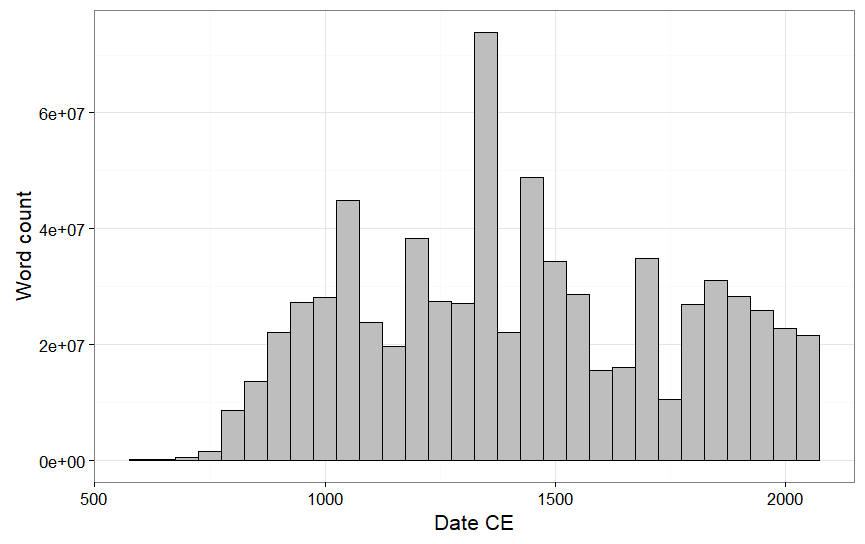}
\label{fig:counts-periods-core-word}
}
\caption{%
Document and word counts per century in the \gls{openiti} corpus versions.
}
\label{fig:counts-periods}
\end{figure}

Table~\ref{tab:data-periodization} shows detailed 
 statistics on the \gls{openiti complete} corpus of texts, including the distribution of texts, sentences,\footnote{Based on the sentence splits created via preprocessing (Section \ref{sec:preprocessing}).} and words over time as reflected in the corpus.  
As illustrated in Figure~\ref{fig:counts-periods}, 
we observe three major jumps in the number of documents: one in the early period, ca, 800 CE, one in the middle period, ca. 1400 CE, and one in the modern period, after 1800 CE. These mostly correspond to increases in the number of words as well.   

As is evident from the statistics, some time periods are more represented in the corpus than others. As in most historical corpora, guaranteeing representativeness is often not possible~\cite{claridge2008historical}. Therefore, we decided to keep all available texts rather than artificially selecting a balanced sub-set of them, in an effort to be as comprehensive as possible. 

\begin{table}[t]
\begin{tabular}{l|rrr||l|rrr}
\toprule
Period (AH) & Texts & Sentences & Words & Period & Texts & Sentences & Words \\
\midrule
\hspace{4.5mm}1--50 	& 43  & 33K 	& 462K & \hspace{1.5mm}751--800 & 407 & 3,980K & 112M \\ 
\hspace{1.5mm}51--100   & 17  & 33K 	& 546K & \hspace{1.5mm}801--850 & 226 & 2,105K & 47M \\ 
101--150 & 48  & 122K	& 3M  & \hspace{1.5mm}851--900 & 240 & 3,396K & 98M \\ 
151--200 & 120 & 549K 	& 11M & \hspace{1.5mm}901--950 & 288 & 2,248K & 55M \\ 
201--250 & 325 & 1,487K & 39M & \hspace{1.5mm}951--1000 & 122 & 1,536K & 45M \\ 
251--300 & 504 & 1,779K & 39M & 1001--1050 & 112 & 858K & 29M \\ 
301--350 & 495 & 2,539K & 65M & 1051--1100 & 103 & 1,533K & 37M \\ 
351--400 & 568 & 2,850K & 59M & 1101--1150 & 82 & 1,713K & 47M \\ 
401--450 & 458 & 2,075K & 46M & 1151--1200 & 81 & 482K & 14M \\ 
451--500 & 481 & 3,350K & 102M & 1201--1250 & 211 & 2,666K & 66M \\ 
501--550 & 266 & 1,919K & 41M & 1251--1300 & 100 & 1,004K & 37M \\ 
551--600 & 443 & 3,392K & 101M & 1301--1350 & 201 & 1,713K & 36M \\ 
601--650 & 291 & 2,361K & 67M & 1351--1400 & 53 & 1,537K & 34M \\ 
651--700 & 313 & 2,216K & 63M & 1401--1450 & 90 & 2,714K & 56M \\ 
\cline{5-8}
701--750 & 456 & 9,597K & 186M &  {\bf Total} & 7144 & 62M & 1,537M \\
\bottomrule
\end{tabular}
\caption{Number of texts, sentences, and words in each 50-year time period in the \gls{openiti complete} corpus. \gls{ah} refers to the Islamic calendar period.
\label{tab:data-periodization}}
\end{table}

\paragraph{Comparison with Shamela}
Initial findings of this work have been reported in the 2016 Workshop on Language Technology Resources and Tools for Digital Humanities (LT4DH)~\cite{belinkov-EtAl:2016:LT4DH}. That work made use of texts from the Shamela text collection. 
This text collection has good coverage of pre-modern texts, and an excellent section of modern texts. However, many foundational texts, particularly texts which are more literary than religious, were absent from that collection. The current 
work is based on the Open Islamicate Texts Initiative, 
which integrates texts from a variety of text collections --- including semi-automated selection of the ideal copy of each text where multiple copies are present. This has the advantage of gathering a greater coverage of the pre-modern era, such that \gls{openiti} has $500$ more texts from the period before 1800 ($15~M$ more words). However, less effort has been made thus far to reintegrate modern texts into \gls{openiti}, so Shamela contains $860$ more modern texts than \gls{openiti}, totaling $77~M$ words. Ongoing initiatives exist to reintegrate the Shamela texts into \gls{openiti} and to increase the number of texts, specifically texts from the early modern era (1800-1900) as that period is when \gls{msa} is said to have started developing.

\subsection{Preprocessing} \label{sec:preprocessing}

We used a combination of OpenNLP%
\footnote{\url{https://opennlp.apache.org}} 
and the \textit{Farasa} toolkit~\cite{N16-3003}%
\footnote{\url{http://farasa.qcri.org}}
to pre-process the corpus. 
Farasa is a popular toolkit for Arabic NLP which was found by multiple studies~\cite{darwish2016farasa} to perform comparably to MADAMIRA~\cite{N16-3003,PASHA14.593.L14-1479,W17-1316}, another popular toolkit. The two toolkits were developed mainly for MSA, but there is no known comprehensive evaluation of them on historical Arabic.\footnote{One study evaluated the two tools on diacritization of Classical Arabic, and found Farasa to perform much better than MADAMIRA~\cite{HamedZesch:2017}.} We provide a small manual evaluation of Farasa on our corpus below, as well as compare the preprocessing results of Farasa and MADAMIRA quantitatively. Our evaluation demonstrates that there is little difference in their performance, and that both are viable choices for preprocessing historical Arabic. 

Next we describe our preprocessing pipeline.%
\footnote{The code to perform this pre-processing is available at \url{https://github.com/albarron/AraProc}.}

\begin{description}
 \item[Sentence splitting.]  
 This is the only component not available in Farasa. Punctuation is inconsistently used in Arabic and a stretch of text between periods may contain many sentences. We used OpenNLP and trained the sentence splitting model on $5~K$ sentences from the AQMAR Arabic Wikipedia Supersense corpus~\cite{Schneider:2012:CLS:2390665.2390726} and NIST's MT06 corpus.%
\footnote{\url{https://www.nist.gov/programs-projects/machine-translation}}
 \item[Morphological segmentation.] 
 The segmenter breaks words into their underlying morphemes. Farasa's segmenter uses SVM$^{rank}$~\cite{joachims2006training} with a linear kernel to determine the best segmentation for each word. 
 \item[Lemmatization.] 
  The lemmatiser is a rule-based system. First, a word is looked up in a word--\{diacritization, lemma\} dictionary and the lemma of the most frequent diacritized word is returned if located. Otherwise, the Farasa segmenter is applied to extract the word's morphemes and the first non-prefix morpheme is looked up again in the dictionary. If found, the mapped lemma is returned; otherwise, the morpheme is returned.
 \item[Part-of-speech tagging.] 
  The POS tagger uses the simplified PATB tagset proposed by~\cite{Darwish:14}. It attempts to find the optimal tag for each morpheme produced by the segmenter, as well as determining the gender (masculine or feminine) and number for nouns and adjectives (singular, dual, or plural). The POS tagger uses SVM$^{Rank}$ to find the best tag for each morpheme as well. 
 \item[Constituency parsing.] %
 This is a re-implementation of the Epic parser~\cite{hall2014less}, which performed best at SPMRL 2013~\cite{bjorkelund2013re}. It uses a conditional random fields  
model trained on features derived from the POS tagger. 
 \end{description}
 We utilize the different output formats produced by the preprocessing throughout this work, except for the parse trees. We still make the trees available to facilitate future syntactic work on the history of Arabic.

\paragraph{Qualitative evaluation}
Farasa is tuned for the news domain and for \gls{msa}; still, ``it can handle other genres along with classical and dialectal Arabic''~\cite{ROMEO2017}. We performed a qualitative analysis of the Farasa results to ensure that it still performed well with Classical Arabic texts. We analyzed the results of the segmentation, POS-tagging, and lemmatization in a set of texts chosen based on genre and date.\footnote{Texts tested were the following (the first digits are the \gls{ah} dates): \\ \texttt{0001NabighaDhubyani.Diwan.JK007511} (pre-Islamic poetry), \\ \texttt{0680IbnSabuni.TakmilaIkmalIkmal.JK000884} (\gls{hadith} collection), \\ \texttt{0685IbnSacidMaghribi.Jughrafiya.Shamela0000463} (geography), \\ \texttt{1405CaliShahrudi.Mustadrakat.Shia002984Vols} (modern devotional). 
} 
In all of the texts, the quality of the Farasa results was high, with relatively few errors in the segmentation and lemmatization, even for words that are obsolete or Classical. In a small sample chosen for manual quantitative analysis, comprising %
a total of 507 words 
in 4 texts, there were 41 mistakes of all kinds, with Farasa correctly segmenting 98.82\%, lemmatizing 98.62\% and POS-tagging 94.28\% of words in this small sample.  
Classical coverage appears adequate --- for example, in a pre-Islamic poem,\footnote{OpenITI document \texttt{0001NabighaDhubyani.Diwan.JK007511-ara1}.} the lemmatizer correctly derived the lemma \transArT{A\$Abp}{mixed group of people} from the plural form \transAr{A\$A\}b}, though this word is almost never used in modern language. This suggests that Farasa's lexicon has an adequate coverage of \gls{ca} vocabulary in spite of being designed for \gls{msa}. 
On the other hand, the POS tagging occasionally produced incorrect gender information, and mislabeled verbs as nouns, though the lemmatization produced the correct form in most cases, and this is an ongoing issue even in \gls{msa} texts. One instance of a specifically \gls{ca} structure that posed a difficulty for Farasa was the energetic form \transArT{l-yltms-n}{he will seek out}, which Farasa treated as a single word in all analyses. The energetic is largely absent from \gls{msa}, and marginal in its use in \gls{ca}. However, Farasa had some difficulty segmenting verbs, especially present-tense verbs, from conjunctions and other prefixes, so this may be a related issue. The output of Farasa is generally useful for a variety of research types, but is specifically useful for concordancing and word frequency counts, since removal of clitics and identification of lemmas poses a significant problem in Arabic computational linguistics. 

\paragraph{Farasa vs.\ MADAMIRA}
Both Farasa and MADAMIRA are high-quality toolkits for Arabic NLP. For periodization, we do not expect to see a large impact of the choice of tool, as we are interested in high-level trends, rather than obtaining a few more points in performance. Nevertheless, we provide here quantitative and qualitative results showing that the differences between the tools are minor and insignificant for our purposes. 

To test this, we ran the two toolkits on 58 texts from our corpus, one from each 25 year time period, and compared their results by computing the character edit distance for each preprocessed sentence. When comparing morphological segmentation, the average edit distance was about 20.1 edit operations, corresponding to 11.2\% of the characters (when dividing by the total sentence length). After accounting for different normalization schemes, the average edit distance was about 8.9 operations, corresponding to just 5.9\% of the characters. 
We also ran a similar evaluation for lemmatization and found even smaller differences between Farasa and MADAMIRA. On lemmatized texts, the average edit distance was 6.8 edit operations, which correspond to 5.4\% of the characters. 


These results indicate that the choice of toolkits for preprocessing would not have a major effect on our analysis. 
More importantly, such small differences do not have a high impact in periodization, where we are concerned with global, overall trends, rather than with improving the state-of-the-art by some small fraction.  

\section{Text Reuse} \label{sec:reuse}
In order to identify instances of text reuse within the corpus, we adapt SKP~\cite{shmidman2016identification}, a recent approach designed  
to efficiently find approximately-parallel passages across a large Hebrew/Aramaic corpus. This method is appealing to use in our case due to the similarity between Arabic, Aramaic and Hebrew: all are languages of Semitic origin with a high morpheme-per-word ratio. First we detail two modifications that we had to apply before running the algorithm in order to handle the nature of the \gls{openiti} corpus. We refer to this enhanced algorithm as SKP-Ar. 
The whole procedure is summarized in Algorithm~\ref{alg:reuse}.

\subsection{Identification of Boilerplate Passages}
Before running the processor-intensive algorithm to identify approximate matches within the corpus, we isolate ``\gls{boilerplate}'' passages that recur verbatim dozens, hundreds, or even thousands of times within the corpus (e.g., Quranic verses are quoted extremely frequently). A common genre in the corpus is  \gls{hadith} collections, Prophetic reports, which must be preceded with a chain of transmission from the person who heard it directly from the Prophet through the main person who transmitted it to the %
latter transmitter. %
Thus, both the chain of transmission and the quotation itself tend to be widely repeated. Others are simply frequently mentioned anecdotes or sayings.


%
To identify \gls{boilerplate} passages, overlapping phrases of length 20 are extracted and counted from the whole corpus (line~\ref{alg:phrase1}) and those phrases which have appeared 25 or more times verbatim are marked (line~\ref{alg:more25}). Phrases are clustered and conflated if they appear overlapped or juxtaposed (allowing for a gap of up to $10$ words between the $20$-gram segments) and the resulting text fragments are marked as boilerplate passages.
These text fragments are ignored by the subsequent stages of the approximate-matching algorithm, allowing them to focus on the more meaningful parts of the text, without getting bogged down in these commonly-recurring exact matches.
 
Appendix~\ref{sec:examples} shows examples of \gls{boilerplate} passages identified in this step. 
%
%
%
%

\begin{algorithm}[t]
\SetAlgoLined\DontPrintSemicolon
\SetKwProg{Fn}{Function}{}{}
\SetKwFunction{Append}{Append}
\SetKwFunction{Conflate}{Conflate}
\SetKwFunction{SortByFrequency}{SortByFrequency}
\SetKwFunction{Top}{Top}
\SetKwFunction{FindBoilerPlates}{FindBoilerPlates}
\SetKwFunction{FindFrequentPhrases}{FindFrequentPhrases}
\SetKwFunction{ConflateFreqPhrasesIntoSingleWords}{ConflateFreqPhrasesIntoSingleWords}
\SetKwFunction{ComputeSkipGrams}{ComputeSkipGrams}
\SetKwFunction{HashSkipGrams}{HashSkipGrams}
\SetKwFunction{GetMatchingHashes}{GetMatchingHashes}
\SetKwFunction{ExtendMatches}{ExtendMatches}
\SetKwFunction{IdentifyReuse}{IdentifyReuse}
\SetKwInOut{Input}{input}\SetKwInOut{Output}{output}

\Input{$Texts$: a list of $T$ documents, ordered chronologically}
\Output{$BoilerPlates$: the boiler plate fragments\\ $ReusedFragments$: the reused fragments }
\Fn{\FindBoilerPlates{$Texts$}}{	\label{alg:boiler}
Initialize dictionary $Phrases(phrase, counter)$\; 
\For{$i \leftarrow 1 $ \KwTo $T$}{	\label{alg:phrase1}
  \For{$j \leftarrow 1 $ \KwTo $J_i \mid |j|=20, step=1$}{
  	$Phrases \leftarrow j$\; 
  }
}
$Phrases \leftarrow [\forall phrase \in Phrases \mid freq(phrase) \geq 25]$\;
$BoilerPlates \leftarrow$ \Conflate{$Phrases$}\;	\label{alg:more25}
\Return{Boilerplates}
}
\BlankLine

\Fn{\FindFrequentPhrases{$Texts$}}{	\label{alg:frequentphrases}
Initialize dictionary $Phrases(phrase, counter)$\; 

\For{$i \leftarrow 1 $ \KwTo $T$}{	\label{alg:frequent1}
  \For{$j \leftarrow 1 $ \KwTo $J_i \mid |j|=4, step=1$}{
  	$Phrases \leftarrow j$\; 
  }
}
$Phrases \leftarrow $ \SortByFrequency{$Phrases$}\;	\label{alg:sortbyfreq}
$ExtremelyFrequentPhrases \leftarrow $ \Top{$Phrases, 35000$}\;	\label{alg:top35k}
\Return{ExtremelyFrequentPhrases}
}
\BlankLine

\Fn{\IdentifyReuse{$Texts$}}{
  $BoilerPlates \leftarrow$ \FindBoilerPlates{$Texts$}\;
  $Texts' \leftarrow [Texts \setminus BoilerPlates]$\;
  $XtremFreqPhrases \leftarrow$ \FindFrequentPhrases{$Texts'$}\;
  $Text'' \leftarrow $  \ConflateFreqPhrasesIntoSingleWords{$Texts', XtremFreqPhrases$}\;
  \BlankLine
  $SkipGrams \leftarrow $ \ComputeSkipGrams{$Texts''$}\; \label{alg:skipgrams}
  $SkipGramsHashes \leftarrow $ \HashSkipGrams{$SkipGrams$}\; \label{alg:hashskipgrams}
  \BlankLine
  Initialize list $ReusedFragments$\;
  \For{$i \leftarrow 1 $ \KwTo $T$}{
    \For{$j \leftarrow 1 $ \KwTo $T$}{
      $BaseMatches \leftarrow $ \GetMatchingHashes{$T_i, T_j, SkipGramHashes$}\; \label{alg:basematches}
      $FullMatches \leftarrow $ \ExtendMatches{$BaseMatches$}\; \label{alg:extendmatches}
      \Append{$ReusedFragments, FullMatches$}\; \label{alg:updatematches}
    }
  }
  \Return{BoilerPlates, ReusedFragments}
  \BlankLine
}
\BlankLine
\caption{The SKP-Ar algorithm for text reuse identification.}
\label{alg:reuse}
\end{algorithm}

\subsection{Identification of Extremely Frequent Short Phrases}
The second step 
involves the identification of frequently recurring $4$-grams. The core of the SKP 
algorithm involves hashing and indexing $4$-grams throughout the whole corpus. This works well for Hebrew and Aramaic, since frequently-recurring Hebrew and Aramaic formulaic phrases tend to be limited to two (sometimes three) words, and thus $4$-grams prove to be effective units in processing the text. However, within the \gls{openiti} corpus, we find many $4$-grams that recur repeatedly. These are often blessings upon the Prophet Muhammad, which are extremely frequent (e.g., \verb|SlY Allh Elyh wslm|, ``peace be upon him'', 
recurs over $5~M$ times).
The prevalence of such phrases could render SKP~\cite{shmidman2016identification} highly ineffective. 

In order to solve this problem, we identify the top $35~K$ frequently-recurring $4$-grams in the corpus (function \texttt{FindFrequentPhrases}). 
We assign each of these phrases a unique 16-bit hash, and henceforth treat each of those phrases as a single word unit, each one with its own unique hash.
When launched on the OpenITI corpus, the selected phrases appeared between 515 and $5~M$ times each.

\subsection{Identification of Approximate Matches}
The function \texttt{IdentifyReuse} details the main steps in the SKP-Ar algorithm. 
As an intermediate step towards finding long parallel passages, we first want to identify all cases of approximately-matching short phrases between the documents. For our purposes, we regard an approximately-matched short phrase as cases of phrases of length five in which four out of the five words are nearly identical. We define a skipgram as any four-word subset of a five-word string. For every 4-out-of-5 skipgram within the text (excluding the boilerplate material, as described above), we assign a 64-bit hash, comprised of the two least-frequent letters of each of the four words (line~\ref{alg:hashskipgrams} in Algorithm~\ref{alg:reuse}). An initial pass of the program reviews the entire corpus and builds a character frequency table of all characters classified as Arabic characters by the Unicode specification. The corpus includes 131 such characters; thus any two-letter combination fits easily into 16-bits, and also allows space for the $35~K$ unique hashes for the frequently-recurring $4$-gram phrases detailed in the previous step. The determination of the two least-frequent letters in a given word is also based upon this initial review of the character inventory within the corpus.
This method conveniently facilitates approximate matches. The use of 4-out-of-5 skipgrams allows passages to match up even though a given word may be subtracted, added, or replaced within the unit. Similarly, the two-letter word hashing allows words to be considered equal despite differences in prefixes, suffixes, or \gls{matreslectionis}.

Now, for any given document (the ``base document''), we tabulate all cases in which one of its skipgram hashes matches a skipgram hash from another document (the ``target document'') (line~\ref{alg:basematches}).  We wish to identify cases in which multiple skipgram matches are in close proximity with one another to form a passage of substantial length. To do so, we generate a two-dimensional graph, wherein each skipgram match is plotted on one axis according to the starting word position in the base text, and on the other axis according to the starting word position in the target text, similar to a dotplot~\cite{Basile:2009,Grozea:11}. We are interested in the cases in which multiple skipgram matches cluster on a more-or-less diagonal line on the graph. To efficiently find such cases, we bin the skipgrams based upon the difference between their two coordinates. We review the bins which contain multiple skipgrams and consider whether those skipgrams can cluster together to form a match containing 16 or more identical words, allowing up to 3 non-matching word positions in between any two matching skipgrams (line~\ref{alg:extendmatches}).

The use of 4-out-of-5 64-bit hashes casts a rather wide net from the start, wherein many identical skipgrams actually point to very different phrases. However, this is compensated by the requirement to have a series of adjacent matching skipgrams. As the number of adjacent skipgrams cluster together, the number of false positives drops progressively lower. In our case, where we require a series of matching skipgrams which match up at a minimum of 16 word positions, we find that the resulting passages are virtually always legitimate cases of text reuse.

\subsection{Results}
We ran the SKP-Ar algorithm on the entire $1.5~G$ word \gls{openiti} corpus. The algorithm first isolated 230,530 unique (though possibly overlapping) frequently-occurring $20$-word phrases (\gls{boilerplate} strings). Each phrase occurs at least $25$ times within the corpus, with the most frequent phrase occuring 4,495 times. In total, we mark 28,491,859  words out of the total $1.5~G$ words as boilerplate text. After eliminating the boilerplate text, the approximate-matching algorithm returned 
$76~M$ pairwise matches, with an average length of $46.7\pm 249$ words per match.
The process took 48 hours, running in parallel on 64 CPUs.

Given that the texts are dated by author's date of death, we allowed some relaxation in finding reused text chunks. 
We counted only cases in which at least $50$ years elapsed between the dates of the earlier and and latter document. 
After this filter, we are left with $57~M$ pairwise matches, with an average length of $31.58\pm 
35.12$.  Note that this filter also eliminates duplicate texts that have the same date.  

Finally, we calculated the total number of reused words within the corpus, leveraging both the set of boilerplate phrases as well as the set of approximately-matching passages. 
After applying the 50-year filter, we were left with a total of 
$292~M$ reused words within the corpus.

Efforts have been carried out to standardize the evaluation of text reuse models~\cite{Potthast:2010col}, but they rely on artificially-generated cases of reuse. Evaluating the performance of a model when applied to a real-life corpus, such as \gls{openiti}, remains a difficult task. Most challenging is evaluating the exhaustiveness of matching. We expect to find extremely important texts to be quoted the most, so the number of matches by text are indicative of whether matches are being correctly identified. Indeed, the largest numbers of \gls{boilerplate} quotations, i.e. quotations repeated 25 times or more, come from the Quran and major works of religious \gls{exegesis} and historical works, all of which we expect to see widely quoted.
Manual examination of the approximate pairwise matches is also positive --- though not quantifiable, rapid scrolling through the lists of results provides easy visual identification of the similarity of the matches, and closer evaluation reveals slight variations in the quotations, often slight reformulation of the phrasing. Visually checking several lists of matches by file did not reveal any obviously flawed matches. 

\section{Automatic Periodization} \label{sec:periodization}

With access to a diachronic corpus, we are able to investigate the linguistic developments which have been claimed to characterize the different stages of \gls{ca}. 
We developed an automatic algorithm for dividing a historical text corpus into time periods. We then applied it to the \gls{openiti} corpus and analyzed the obtained results.\footnote{The periodization code is available at \url{https://github.com/boknilev/periodization}.}

\subsection{Word-Embedding-based Neighbor Clustering}

Given the nature of language change, we note that the language in two consecutive time periods should in principle be more similar than the language in two remote time periods. Therefore, we apply a chronologically-constrained hierarchical clustering algorithm that is only allowed to merge consecutive time periods. The core of the algorithm is based on a word-embedding function and a distance function. The word-embedding function takes a text document and generates a word-embedding matrix. The distance function takes two word-embedding matrices, corresponding to two time periods, and computes the distance between them. Below we discuss specific instantiations of these functions.
Our algorithm can be seen as a word-embedding-based variant of the Variability-based Neighbor Clustering (\gls{vnc}) algorithm for periodization~\cite{GriesHilpert}, where we replace the measure of variability by a distance measure based on word-embedding matrices. Word embeddings are attractive to use for this purpose because they provide a soft notion of language use, with similar words having similar vectors in the word embedding space~\cite{Mikolov-Yih-ZWeig:2013}.  
We name our periodization algorithm Word-Embedding-based Neighbor Clustering (\gls{wenc}). 

We assume a collection of texts, $\mathcal{T}$, with known dates. We first bin the texts into initial time periods $\mathcal{P} = \{P_1, ..., P_{|\mathcal{P}|}\}$, that are ordered chronologically (e.g., centuries). 
That is, for each $i<j$, all the texts in $P_i$ are dated earlier than all the texts in $P_j$. 
The texts in each time period are concatenated into documents that are input to the periodization algorithm (see Algorithm~\ref{alg:periodization}). 
We start by training initial word embedding models (line~\ref{alg:init-models}). Then, we iteratively look for the next best possible merge of time periods until there are no more time periods to merge (line~\ref{alg:iter-best-merge}). At each iteration we record the best merges and distances (lines~\ref{alg:update-pairs}-\ref{alg:update-dists}).  

The algorithm utilizes a function \texttt{FindBestMerge} that takes a collection of documents and their corresponding word embedding models and computes the distances between each consecutive pair of documents (line~\ref{alg:distances}). It then finds the best pair (line~\ref{alg:argmin}), concatenates the two documents (line~\ref{alg:merge-text}), trains a word embedding model on the new concatenated document (line~\ref{alg:merge-model}), and updates the list of texts and models (lines~\ref{alg:update-texts}-\ref{alg:update-models}). 

\begin{algorithm}[t]
\SetAlgoLined\DontPrintSemicolon
\SetKwProg{Fn}{Function}{}{}
\SetKwFunction{Periodize}{Periodize}
\SetKwFunction{TrainWordEmbModel}{TrainWordEmbModel}
\SetKwFunction{FindBestMerge}{FindBestMerge}
\SetKwFunction{ComputeDistance}{ComputeDistance}
\SetKwFunction{Append}{Append}\SetKwFunction{Concat}{Concat}
\SetKwFunction{ArgMin}{ArgMin}\SetKwFunction{UpdateList}{UpdateList}
\SetKwInOut{Input}{input}\SetKwInOut{Output}{output}

\Input{$Texts$: a list of $T$ documents, ordered chronologically}
\Output{$MergedPairs$: the merged clusters\\ $MergedDistances$: the corresponding distances}
\Fn{\Periodize{$Texts$}}{
Initialize list $Models$\; 
\For{$i \leftarrow 1 $ \KwTo $T$}{
  $Models[i]$ $\leftarrow$ \TrainWordEmbModel{$Texts[i]$}\; \label{alg:init-models}
}
\BlankLine

Initialize lists $MergedPairs$, $MergedDistances$\;
\While{$|Texts| > 1$}{
  $BestPair$, $BestDist$, $Texts$, $Models$ $\leftarrow$ \FindBestMerge{$Texts$, $Models$}\; \label{alg:iter-best-merge}
  \Append{$MergedPairs$, $BestPair$}\; \label{alg:update-pairs}
  \Append{$MergedDistances$, $BestDistance$}\; \label{alg:update-dists}
}
\Return{$MergedPairs$, $MergedDistances$}
}
\BlankLine
\Fn{\FindBestMerge{$Texts$, $Models$}}{
  Initialize list $Distances$\;
  \For{$i \leftarrow 1 $ \KwTo $|Texts| - 1$}{
    $Distances[i] \leftarrow$ \ComputeDistance{$Texts[i]$, $Texts[i+1]$}\; \label{alg:distances}
   }
  $BestPair$, $BestDistance$ $\leftarrow$ \ArgMin{$Distances$}\; \label{alg:argmin}
  $MergedText \leftarrow$ \Concat{$Texts[BestPair]$}\; \label{alg:merge-text}
  $MergedModel \leftarrow$ \TrainWordEmbModel{$MergedText$}\; \label{alg:merge-model}
  $Texts \leftarrow$ \UpdateList{$Texts$, $MergedText$}\; \label{alg:update-texts}
  $Models \leftarrow$ \UpdateList{$Models$, $MergedModel$}\; \label{alg:update-models}
  \Return{$BestPair$, $BestDistance$, $Texts$, $Models$}
}
\BlankLine
\caption{\gls{wenc}: word-embedding-based neighbor clustering for automatic periodization. 
}
\label{alg:periodization}
\end{algorithm}

\subsubsection{Word Embeddings and a Distance Measure}

The periodization algorithm relies on training word embeddings on texts in each time period (line~\ref{alg:distances} in Algorithm~\ref{alg:periodization}). We obtain the word embeddings using \texttt{Word2Vec}~\cite{DBLP:journals/corr/abs-1301-3781,DBLP:journals/corr/MikolovSCCD13,DBLP:conf/naacl/MikolovYZ13} as implemented in \texttt{gensim} \cite{rehurek_lrec}. Specifically, we train the CBOW algorithm with negative sampling and the following default settings defined in \texttt{gensim}: word embedding dimensionality of 100, 5 negative samples, and a window size of 5 words.   

Given two word embedding matrices, $W_1$ and $W_2$, trained on different corpora, we need to define a distance measure between them. One option could be to directly calculate the distance with respect to some norm: 
\begin{equation}
\texttt{ComputeDistance}(W_1, W_2) = ||W_1 - W_2||
\end{equation}
However, since the two word embedding matrices are trained independently, we have no guarantee that this distance measure would yield meaningful results. Moreover, the stochastic nature of the word embedding training algorithm precludes a direct comparison of words from different embedding models.  
To avoid this problem, we follow~\cite{P16-1141} and  align the two matrices using orthogonal Procrustes:
\begin{equation}
\texttt{ComputeDistance}(W_1, W_2) = \min_{Q: Q^TQ = I} ||QW_1 - W_2||_F
\end{equation}
where $||\cdot||_F$ is the Frobenius norm of matrix $\cdot$ and $Q$ is an orthogonal matrix that rotates the word embedding matrix $W_1$ towards $W_2$. The solution to this minimization problem is given by the best rotation matrix and can be found with singular value decomposition (\gls{svd})~\cite{Schonemann1966}:
\begin{equation}
R = \argmin_{Q: Q^TQ = I} ||QW_1 - W_2||_F = UV^T
\end{equation}
where $W_2 W_1^T = U \Sigma V^T$ is the \gls{svd} decomposition.

\subsection{Experiments}
For the periodization experiments, we consider two possible initial time divisions: 50 and 100 year bins.\footnote{Since the \gls{openiti} corpus is dated by Islamic years, we use 50/100 Islamic year bins.} Table~\ref{tab:data-periodization} shows the number of texts, sentences, and words, for each 50-year bin. The 100-year bins are simply a concatenation of each two consecutive 50-year bins. We find that the very early time periods contain much less text, so we merge the bins for the first 200 years in all of the experiments (2 and 4 bins are merged in the case of the 100-year and 50-year bins, respectively). 
In the following, we investigate the effect of preprocessing  on periodization, by  
considering two input formats for the periodization algorithm: plain text 
and lemmas. Working with lemmas allows the periodization algorithm to focus more on lexical properties rather than surface forms. However, the morphological 
lemmatization performed by Farasa (Section~~\ref{sec:preprocessing}) 
is an automatic process that may produce errors, so we also run the periodization algorithm on plain text. 

We also investigate the effect of text reuse by comparing the periodization results on the full corpus to running the same algorithm on a version of the corpus where reused text chunks were removed. 

In all cases,
we run the \gls{wenc} periodization algorithm (Algorithm~\ref{alg:periodization}), record the hierarchical merges and their distances, and plot the results in dendrograms. 

\bigskip

\begin{figure}[t]
\subfloat[100-year bins, plain.]{
\centering
\includegraphics[width=0.5\linewidth]{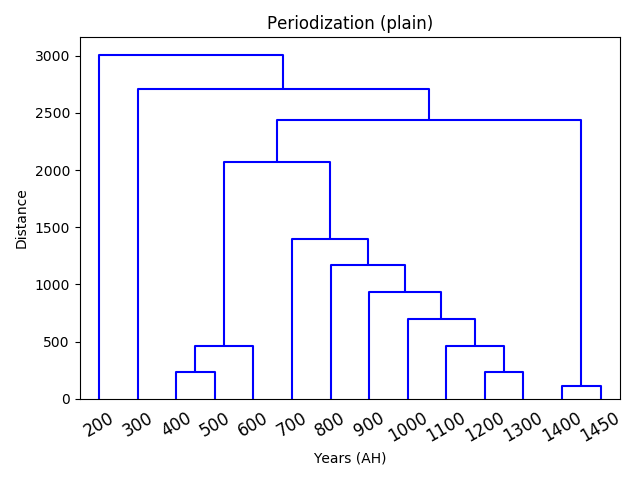}
\label{fig:dendrogram-100-unsegmented}
}
\subfloat[100-year bins, lemmatized.]{
\centering
\includegraphics[width=0.5\linewidth]{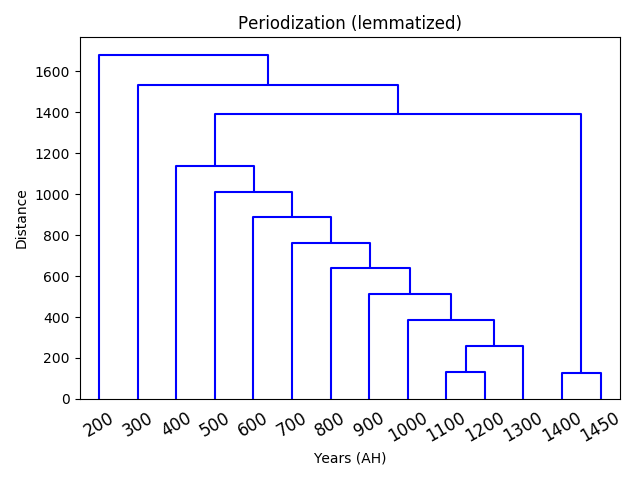}
\label{fig:dendrogram-100-lemmatized}
}\\
\subfloat[50-year bins, plain.]{
\centering
\includegraphics[width=0.5\linewidth]{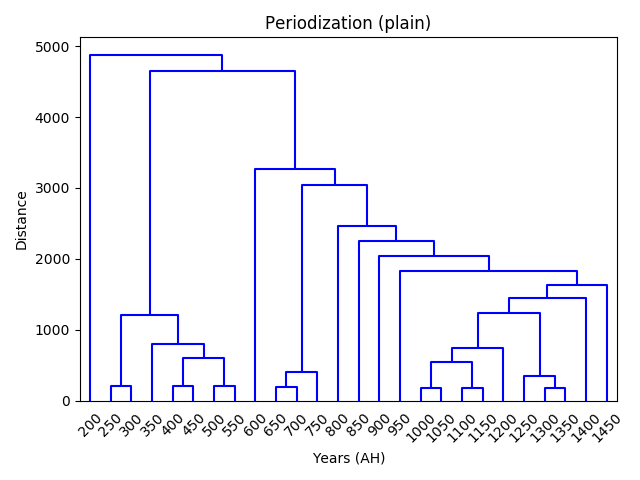}
\label{fig:dendrogram-50-unsegmented}
}
\subfloat[50-year bins, lemmatized.]{
\centering
\includegraphics[width=0.5\linewidth]{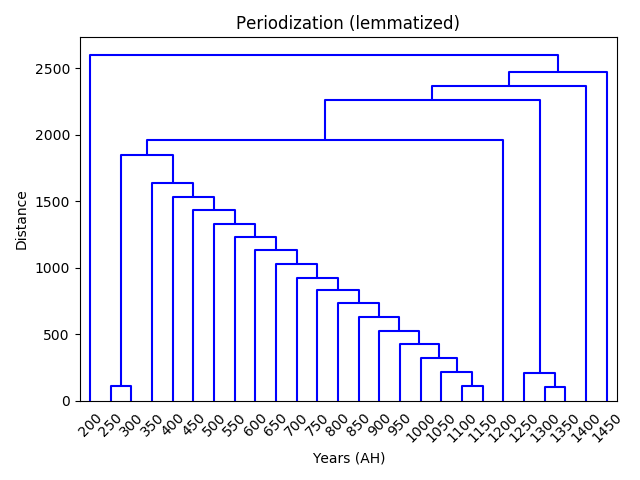}
\label{fig:dendrogram-50-lemmatized}
}
\caption{Periodization on 50-year and 100-year bins, using plain 
and lemmatized text. The y-axis shows the cumulative distance between merged clusters. 
}
\label{fig:dendrograms-100}
\end{figure}

Figures~\ref{fig:dendrogram-100-unsegmented} and~\ref{fig:dendrogram-100-lemmatized} show the results of running \gls{wenc} 
on 100-year time periods, using plain and lemmatized texts, respectively. In both cases, we see a split into three main periods: early period until 200/300 \gls{ah}, middle period from 200/300 \gls{ah} to around 1300 \gls{ah}, and a late period from that time to modern days. 
In the case of 50-year bins (Figures~\ref{fig:dendrogram-50-unsegmented} and~\ref{fig:dendrogram-50-lemmatized}), we can observe a more fine-grained periodization. The two figures are very different: the periodization based on plain texts leads again to three large time periods, more or less corresponding to the results with 100-year bins. The periodization based on lemmatized texts exhibits a very large middle period, with comparatively shorter early and late periods. The reason may be that some of the differences between time periods are not as strong when abstracting over the word forms and working with lemmas. 
Interestingly, the algorithm does not always show the split between texts before and after 1300 AH (1882 CE) that should represent the \gls{ca} and \gls{msa} boundary as it is normally portrayed in the literature. However, all but Figure \ref{fig:dendrogram-50-unsegmented} seem to point towards a differentiation between a modern and pre-modern period. %

The algorithm should in principle abstract over genre effects by comparing only shared word embeddings between time periods. However, word embeddings are trained based on their contexts which in turn are influenced by genre. Therefore it should be noted that a consistent grouping which splits the earliest bin from later bins could be due in part to genre effects.  The 200 year bin includes all texts from the 1st century, which is dominated by poetic collections. Texts which have the title ``\gls{diwan}'', meaning a poetic collection (some poetic collections may have other titles), are primarily found in the first several centuries. 
Of the 148 distinct works in the first two centuries, 67 or 44\% are \glspl{diwan}, but by the year 300, only 14\% of distinct works are \glspl{diwan} and only 3.4\% of all books in the corpus are \glspl{diwan}. Most of the corpus is prose and may differ significantly from poetry in style and use of words in specific contexts. In interpreting the results of the periodization, we should be cautious due to the interaction between genre and text date in the corpus. 

\subsubsection{Text Reuse and Periodization}
\begin{figure}[t]
\subfloat[100-year bins, ``hollowed'']{
\centering
\includegraphics[width=0.5\linewidth]{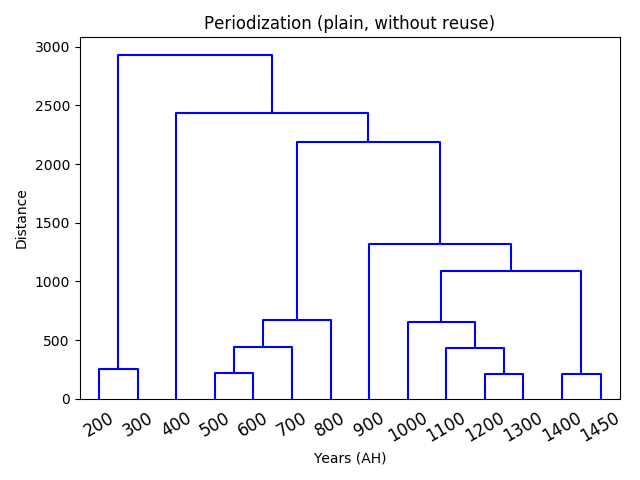}
\label{fig:dendrogram-100-unsegmented-noreuse}
}
\subfloat[50-year bins, ``hollowed'']{
\centering
\includegraphics[width=0.5\linewidth]{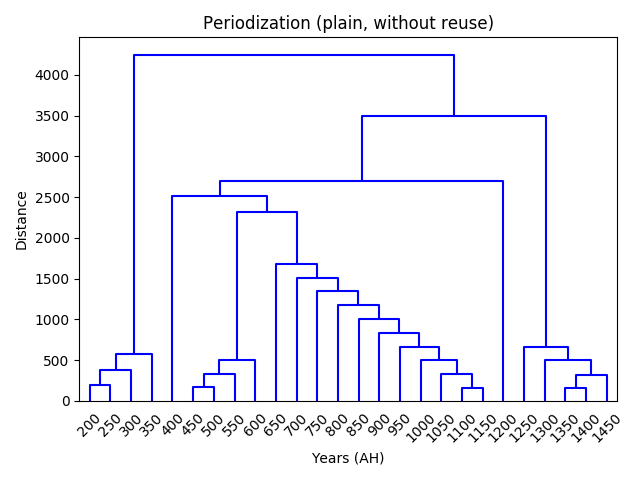}
\label{fig:dendrogram-50-unsegmented-noreuse}
}
\caption{Periodization on 50-year and 100-year bins after removing reused text.}
\label{fig:dendrograms-noreuse}
\end{figure}

The detection of periods of language change can be affected by quotation and general reuse of text chunks from early periods. The text reuse algorithm detects such texts and so we use it here to remove all cases of reuse. The ``un-reused'' or ``hollowed'' corpus is obtained by removing, for each detected match, all later cases of reuse, while keeping the earliest instance. 
Figure~\ref{fig:dendrograms-noreuse} shows the resuts of running \gls{wenc} on the hollowed corpus in 50-year and 100-year bins (showing the plain text version). The results are fairly similar to the periodization results on the full corpus (Figure~\ref{fig:dendrograms-100}), but there appears to be a clearer separation around 800-900AH, especially in the 100-year bins case. Thus, after removing instances of text reuse, we find a clearer division into different pre-modern time periods. 

The separation of the modern period from the pre-modern period becomes more evident once textual quotations are removed, with the 1400s CE (1979 CE to present) clearly separated, and in Figure \ref{fig:dendrogram-50-unsegmented-noreuse}, the more expected differentiation of 1250 AH (1770 CE) separating from the previous period. This is right at the beginning of the period in which \gls{msa} developed. With 100-year bins, 1400 AH would represent authors who died between 1300 AH (1882 CE) and 1979 CE, i.e. a group born around the 1820s to the early 20th century. These represent the second or third generation of modernizing authors. That this would become somewhat more evident in the ``hollowed'' corpus is logical, as the texts in the corpus are inherently conservative and religious in nature, quoting heavily from earlier works. Removing those quotations makes the modern sections of these texts more salient to the algorithm.   

\medskip
Overall the automatic periodization separates out the earliest  and latest texts (200-300 and 1400-1450, sometimes including 1350) from a core bin that occupies the periods roughly between 400 AH (1009 CE) and 1300 AH (1882 CE). The first bin appears to correspond to Fischer's \gls{psca} era, containing a large quantity of pre-Islamic and early Islamic poetic texts, while the core bin is \gls{ca} proper. \gls{msa} comes surprisingly late, even assuming a 70-year author lifespan. There is some support in the periodizations for the claimed change in the language due to the end of the Abbasid empire in 1258 CE (656 AH), with a cluster break clearest in the ``hollowed'' periodizations. A surprising result which requires further research is the consistent break around 900 AH (1494 CE). Though the literature has not considered this as a transition period, it does corresponds to the end of the Islamic state in the Iberian peninsula (1492 CE), and the rise of the Ottoman empire (conquest of Syria and Egypt during 1510s CE).

\section{Expert Periodization of Arabic} \label{sec:expert}
The corpus also supports less automated approaches to periodization that still rely on computational tools. These approaches are particularly useful for assessing the validity of periodizations based on impressionistic analyses of Arabic from earlier publications. 

\subsection{The Lifespan of Arabic Words} \label{sec:lifespan}

It is possible that the impression of \gls{sa} as unchanging is actually a myth rather than reality, that \gls{sa} actually changes just as quickly as other written languages. In order to investigate this question we use the corpus to check whether there is a quantitative difference in the development of Arabic writing and other languages for which historical corpora are available. 

To do this, we track the ``lifespan'' of Arabic and English words in two corpora: \gls{openiti core} and the Corpus of Historical American English (\gls{coha})~\cite{davies2010corpus}. We used lemmas rather than words in both cases. For \gls{openiti}, we use the lemmatization provided by Farasa, but since Farasa does not reject non-words, we use MADAMIRA~\cite{PASHA14.593.L14-1479} to discard lemmas that are not actually Arabic words (incorrectly spelled words, etc.).  For every lemma in the corpus, we find its first and last chronological usages. We discard words which occur only once or in a single year, which may be misspellings (often the case in \gls{coha}), or which have apparently short lifespans since they occur in the very last year/decade (in COHA) in the corpus.

\begin{figure}[t]
\centering
\captionsetup[subfigure]{position=b}
\subfloat[\gls{openiti core}.]{
\includegraphics[width=0.5\textwidth]{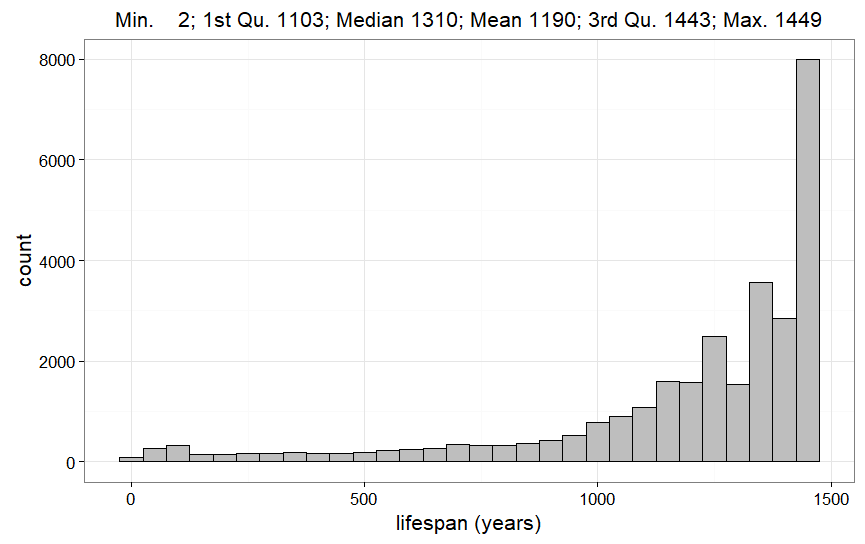}
\label{fig:arabic-lifespan}
}
\subfloat[\gls{openiti core} ``hollowed'']{
\includegraphics[width=0.5\textwidth]{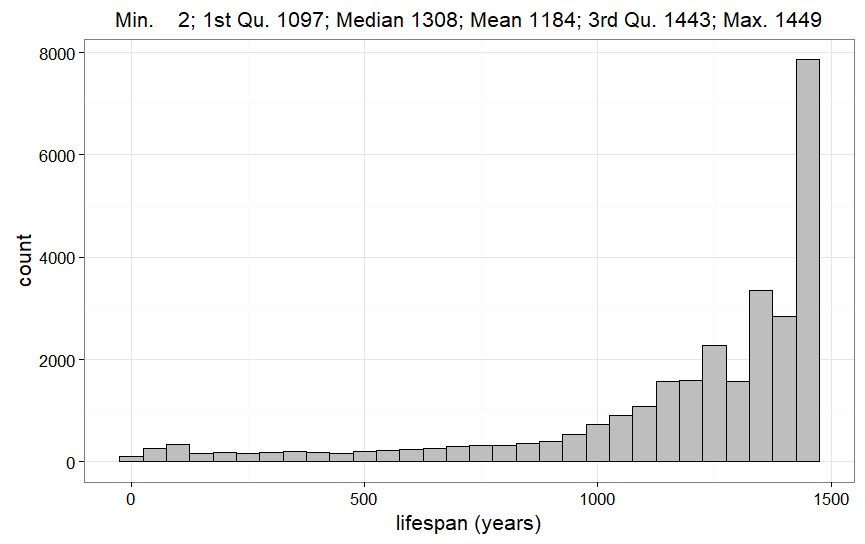}
\label{fig:lifespan-ar-core-hollow}
} \\
\subfloat[\gls{openiti core} 1225AH-1450AH (1810CE-Present)]{
\includegraphics[width=0.5\textwidth]{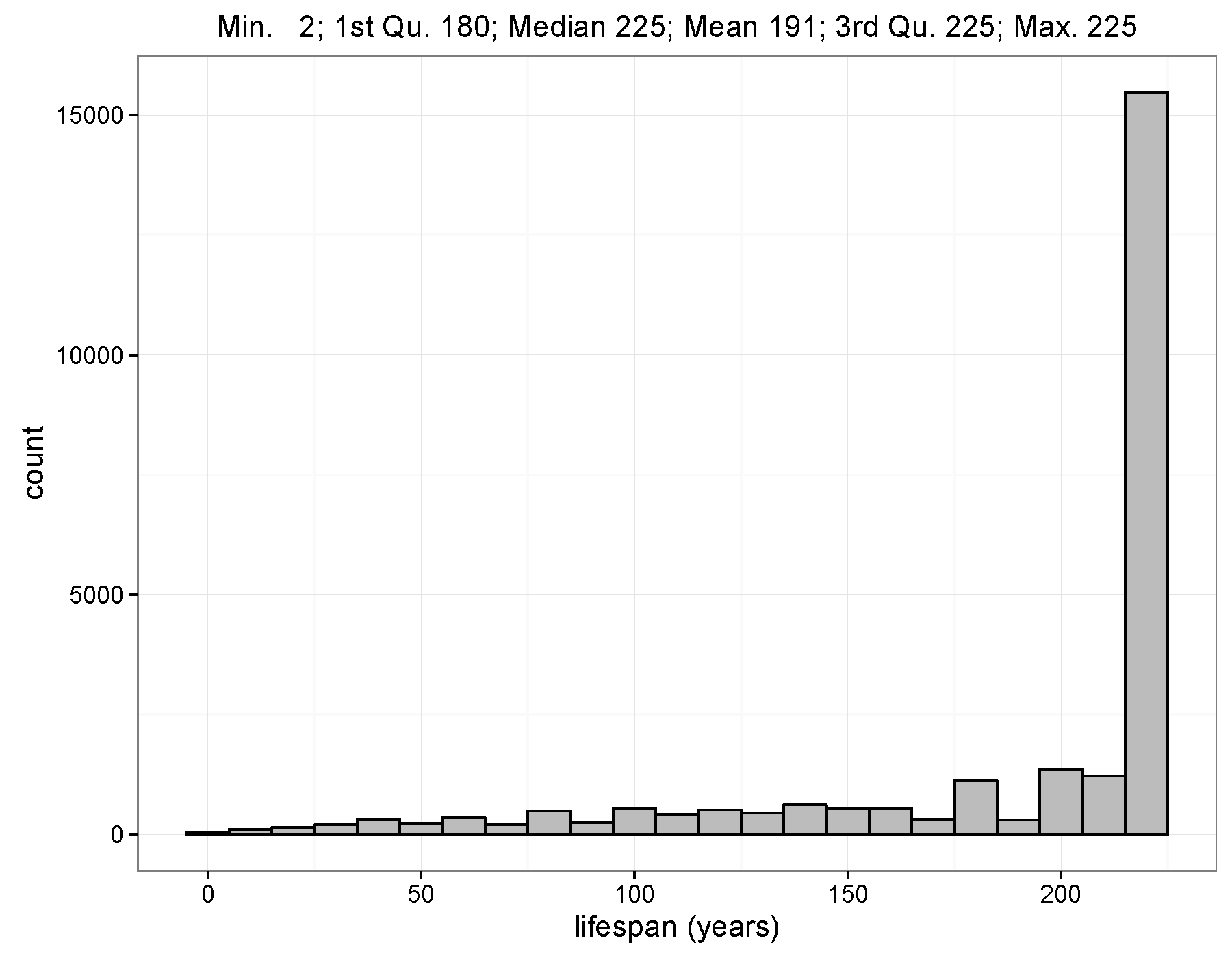}
\label{fig:lifespan-ar-lastbin}
}
\subfloat[\gls{coha}.]{
\includegraphics[width=0.5\textwidth]{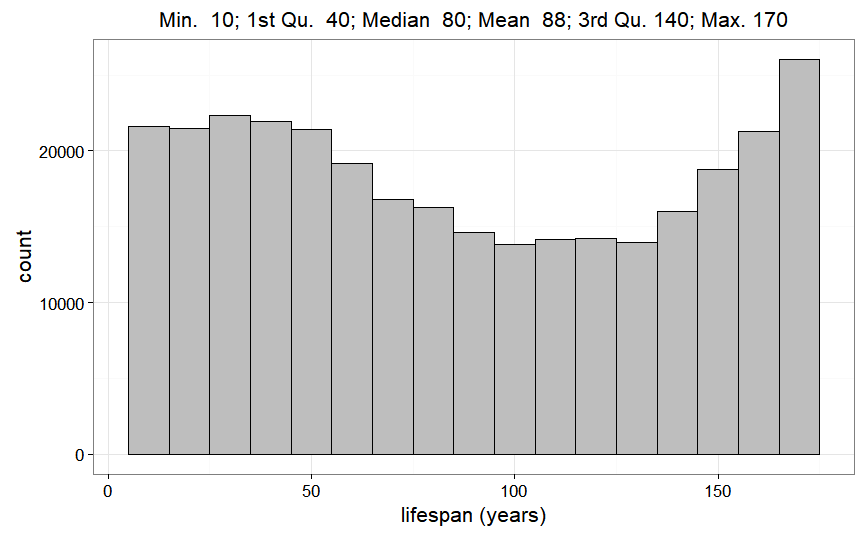}
\label{fig:english-lifespan}
}
\caption{Distribution of word lifespans in Arabic, using (a) \gls{openiti core}, (b) a hollowed version without text reuse, (c) the last 225 years, and (d) the \gls{coha} English corpus. 
}
\label{fig:lifespans-all}
\end{figure}

Figure~\ref{fig:lifespans-all} shows the difference between Arabic and English word lifespans. The comparison reveals that Arabic words tend to have a very long life span in comparison with English words.   In OpenITI the average Arabic word lifespan is  1,190  years (see figures for further descriptive statistics), approximately $83\%$ of the time span of the entire corpus. In \gls{coha}, the average English word lifespan is 88 years, about $45\%$ of the overall time span of the 190 year corpus. Of course, COHA is a much narrower corpus, only covering 200 years, from 1810-2000s.  To control for the 200 year span of COHA versus the much longer span of our corpus, we also ran the same analysis on the set of all thirteen 200 year spans of the corpus (in 100-year increments, with one 225 years span for the final bucket). All lifespan analyses exhibited a similar trend, with long left-tailed distributions;  Figure~\ref{fig:lifespan-ar-lastbin} shows an example for the last time span. Morover, under this analysis, the mean lifespan of words was 161 years, nearly double COHA's average word lifespan. These results confirm that Arabic words tend to have longer lifespans than English ones. 

Arabic words do not disappear as quickly from the lexicon as English words do, suggesting that indeed Arabic does change less quickly than English. This provides quantitative confirmation of an earlier qualitative observation: while Arabic-speaking schoolchildren can read a text from 750 CE, an English-speaking middle school student would have a much more difficult time reading Beowulf, an Old English poem from ca. 1000 CE. It also means that any measures of change in Arabic will need to be more sensitive than measures used for English in order to produce a meaningful result. 

It is possible that the apparently long lifespan of these words is due to extensive quotation of texts including archaic words. In a sense, this does not really matter from the perspective of a language user, since they still must be able to understand the quotation regardless (footnotes may be provided to help with this in some texts, but not all.) To determine the extent that quotation influences word lifespan in Arabic, we ran the same lifespan measurements on the ``hollowed'' version of the lemmatized corpus with reuse removed. The results are shown in Figure \ref{fig:lifespan-ar-core-hollow}, and differ very little from the results on the normal lemmatized corpus. This shows that the long lifespans of Arabic words are clearly not the result of quotation alone.






\subsection{New Words Over Time}
Another claim in the literature is that the number of new words increased significantly during the development of \gls{msa}, as new terms were needed to refer to 
European technology and ideas suddenly becoming available in the Arab world due to colonization and modernization efforts~\cite{Newman13MSAOxfordHandbook}. We can use the corpus to confirm that this is indeed accurate, and to investigate whether there are other periods when an increase occurred in the number of new vocabulary items. 

\begin{figure}[t]
\captionsetup[subfigure]{position=b}
\subfloat[\gls{openiti}.]{
\includegraphics[width=0.5\textwidth]{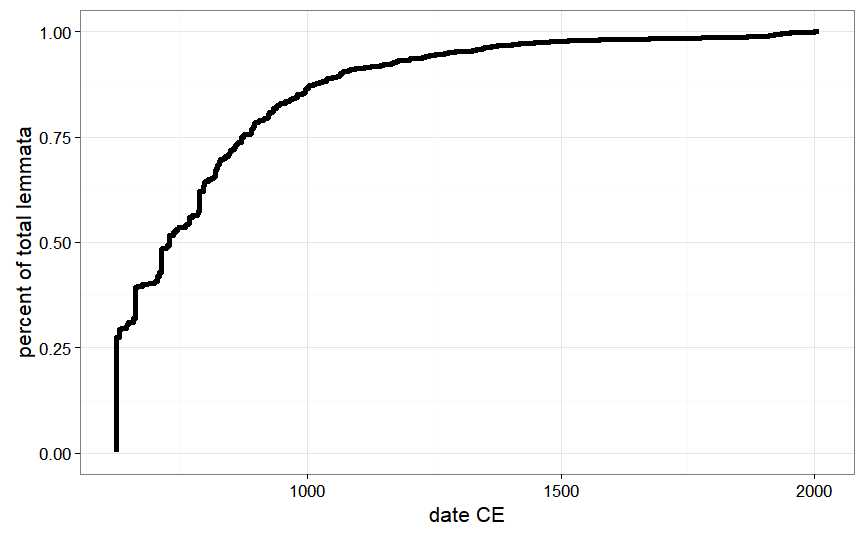}
\label{fig:newlemmasall}
}
\captionsetup[subfigure]{position=b}
\subfloat[\gls{coha}.]{
\includegraphics[width=0.5\textwidth]{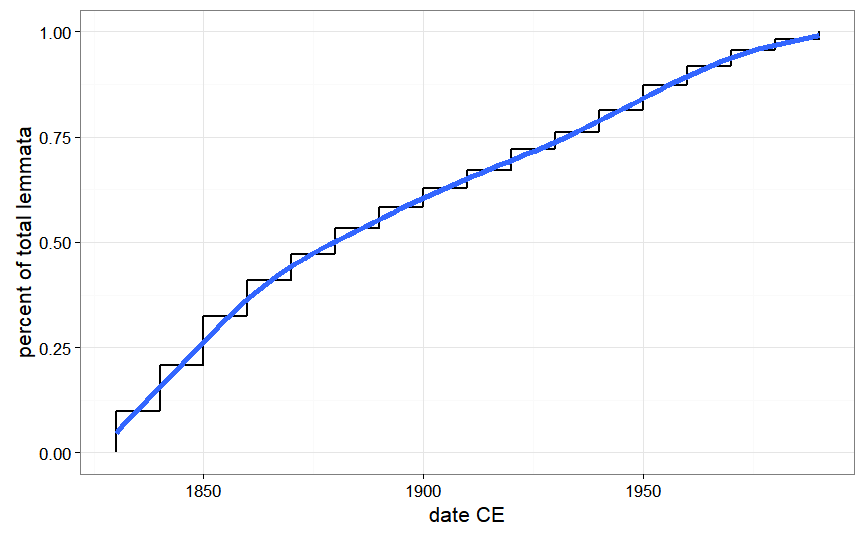}
\label{fig:newlemmasE}
}
\captionsetup[subfigure]{position=b}
\subfloat[\gls{openiti}, modern period.]{
\includegraphics[width=0.5\textwidth]{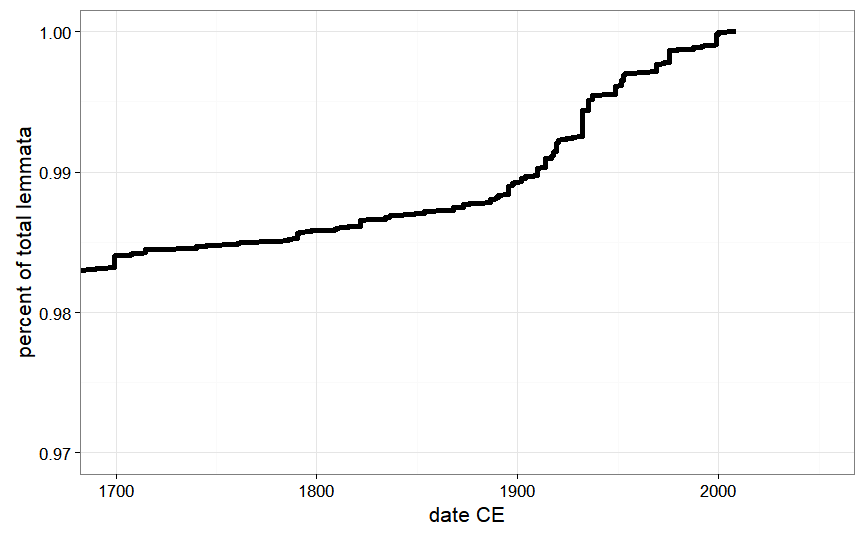}
\label{fig:newlemmas1700}
}
\captionsetup[subfigure]{position=b}
\subfloat[\gls{openiti}, medieval period.]{
\includegraphics[width=0.5\textwidth]{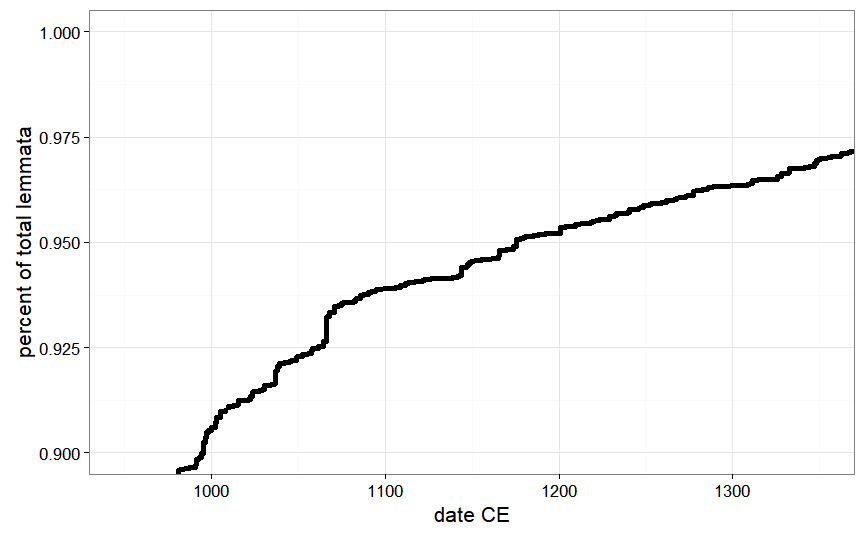}
\label{fig:newlemmas1000}
}
\caption{New lemmas as a percentage of total lemmas over time in Arabic in \gls{openiti} and English in \gls{coha}.}
\label{fig:newlemmas}
\end{figure}

Figure \ref{fig:newlemmasall} shows that, as expected, new lemmas in Arabic developed very rapidly, a corollary of the long lifespan of Arabic words. Compare this to the English results which show a steadier rise (Figure \ref{fig:newlemmasE}). Zooming into the modern era in Figure \ref{fig:newlemmas1700}, we see that fully $1\%$ 
of Arabic words were added just in the 20th century (really earlier, since these are date-of-death measures), a much more rapid increase than in the previous two centuries, which added slightly more than $0.5\%$ 
of lemmas to the lexicon. There appears to be another period of rapid growth at the beginning of the second millennium, with a large jump in vocabulary between 1000 and 1100, and again a jump starting in 1150, but they are not as clear as the increase in the modern period (Figure \ref{fig:newlemmas1000}). Some caution is needed as a single author or work could easily cause these jumps in the vocabulary, but this is an interesting result: the account of Arabic history that claims a decline in the language following the end of the \gls{abbasid} empire would predict that vocabulary addition would level at this point, rather than increasing. On the other hand, the breakdown of central authority might decrease standardization of vocabulary and increase diversity in specialized terminology.   %

\subsection{Verifying Previous Periodizations}

Pre-Standardized Classical Arabic (\gls{psca}) does have distinguishing linguistic features, though these are largely found in the Quran or in rare poetic attestations. Moreover, most of these are slight differences in assimilation or vocalization, and would not show up in the written text in an easily distinguishable way. One of the few testable claims is that prior to the 8th century CE, the formation of abstract conceptual nouns was done via a phrase, \transArT{ElY jhp Al-}{from the perspective of}, but was replaced with a suffix \transArT{-yp}{-ity, -ness} in the period of Standardized Classical Arabic (\gls{sca})~\cite{Ali87Vocab}.
Two suggested phrases based on the \transArT{ElY jhp Al-}{from the perspective of} structure are \transArT{ElY jhp Al-xyr}{charity, goodness} and   \transArT{ElY jhp Al-Edl}{justice, fairness}. A concordance search on the lemmatized corpus finds these structures are basically unattested in the data with a total of less than 30 attestations, all of which post-date the 8th century CE. Nor is this structure attested in Arabic papyri via the Arabic Papyrology Database.\footnote{\url{http://www.apd.gwi.uni-muenchen.de:8080/apd/project.jsp}.} It is therefore unclear where this claim originates since it is so poorly supported by the data.\footnote{The source of this reference,~\cite{Ali87Vocab}, cites a work that happens not to be in his bibliography and is therefore difficult to locate.} 

More testable features are claimed for the break between \gls{sca} and \gls{pca}. These include the use of the adverb \transArT{AyDA}{also}\footnote{\cite{EALLClassicalArabic} also suggests the \gls{pca} adverb  \transArT{xASp}{especially}, e.g. \transAr{x\={a}\d{s}\d{s}at-an} but this is homographic with the word \transArT{x\={a}\d{s}\d{s}a}{the elite} and so cannot be easily distinguished from one another using computational methods.}, and the development of adjectives that show a suffix \transAr{-Any} such as \transArT{jsmAny}{bodily} and \transArT{rwHAny}{spiritual}~\cite{EALLClassicalArabic}. These are relatively insignificant changes, but at  least they can now be verified or disproved using OpenITI.

\begin{figure}[t]
\centering
\includegraphics[width=.5\linewidth]{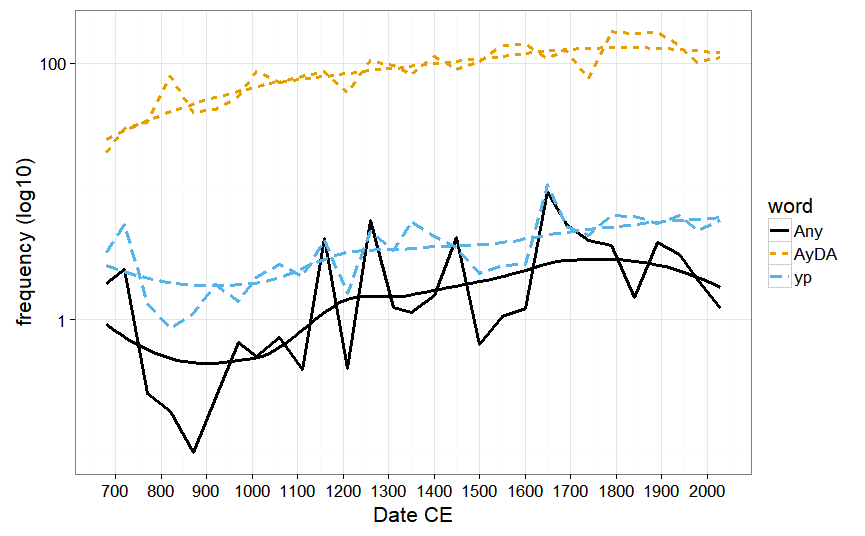}
\caption{Relative frequencies of words and suffixes, with actual frequencies and \gls{loess} smoothed lines.}
\label{fig:freqs}
\end{figure}

Using the segmentation produced by Farasa to remove extraneous clitics (Section~\ref{sec:preprocessing}), we are able to investigate whether these claims are accurate. Figure \ref{fig:freqs} shows the relative frequencies for the word \transAr{AyDA}, and the combined frequencies for the words \transAr{jsmAny} and \transAr{rwHAny}, as well as  
several other words suggested in the literature, which use an abstracting suffix \transAr{-yp}.\footnote{The words are \transArT{mEqwlyp}{intelligibility},\transArT{Eqlyp}{mentality}, \transArT{nHwyp}{specificity}, \transArT{xyryp}{charitability}, and \transArT{Edlyp}{justice} all suggested by \cite{Ali87Vocab}.}
For the word \transAr{AyDA}, there is no relationship to different eras, with this word functioning as an adverb even in very early texts. In OpenITI our earliest clear attestation is in a text with author DOD of 68 HA (688 CE)\footnote{OpenITI document \texttt{0068AbdAllahIbnCabbas.GharibQuran.Shamela0023622}. We find an even earlier attestation in the Arabic Papyrology database, being used as `also' in papyrus ``P.StoetzerSteuerquittungen 2'' from 677 CE.} It increases in usage over time, but shows no periods of significantly greater growth. It is worth noting that its use was judged negatively, with Ibn al-Sikkit's (d. 858 CE) dictionary providing precise rejoinders to mock anyone who uses the word in its adverbial meaning as `also'~\cite{LaneDict}. This establishes that the `also' meaning is attested at least by the 9th century, but also that this term was undergoing some form of change for it to be subject to meta-linguistic judgment. However, this significantly predates the claimed \gls{pca} development date, nor is there a discernible increase in its frequency at that time. 

The other two variables do correspond roughly to the claimed \gls{pca} era, but increases in their usage actually seem to precede the Mongol conquests in 1258 CE --- since all figures here are dates of death,
clearly use of these suffixes was on the rise well before that time, and recent research suggests that the Mongols simply finished off a process of decline already underway in Iraq and Iran during that era~\cite{Romanov17Biographical}. The two variables 
do show remarkable similarity in their frequencies over time, suggesting that the increase in their usage does reflect a change in linguistic eras, even if it begins prior to 1258 CE.

\section{Conclusions} \label{sec:conclusion}
The Arabic language has a long and diverse history spanning more than 1400 years. The written Arabic language, Standard Arabic (\gls{sa}), is often thought to be a more or less monolithic language with little change before the modern period. In this work, we process \gls{openiti}, a large-scale diachronic corpus of Arabic, and investigate the question of periodization in different ways. We identify instances of text reuse in the corpus, develop an automatic periodization algorithm, and investigate existing claims about the periodization of Arabic. We find that although words do persist relatively longer in Arabic than in English, there is evidence for several distinct periods in the language's development.

\gls{openiti} represents the largest publicly available diachronic corpus of Arabic to date. We preprocessed the corpus to make it more amenable to natural language processing, and the results of this are also available for use. The nature of the corpus is such that texts frequently quote from one another, and using an efficient algorithm we were able to identify $292~M$ words of reused text, nearly $20\%$ of the total corpus. We were able to produce a ``hollowed'' version of the corpus with this data removed, both for the plain-text and preprocessed versions of the corpus. 

This corpus and the tools we develop allow us to answer open questions about the history of Arabic. The corpus allows us to establish that Arabic vocabulary does indeed change more slowly over time than in English.  The automatic periodization algorithm we develop confirms established periodizations of Arabic, while suggesting new ones.  It shows that the oldest periods of Arabic and the most modern ones are both separate from a core period  stretching from approximately 400 CE (1009 CE) until 1300 AH (1882 CE), reflecting the prototypical Classical Arabic (\gls{ca}). The data from the automatic periodization and from the evidence of new words in Arabic both strongly support established periodizations that divide Modern Standard Arabic (\gls{msa}) from \gls{ca}. Both automatic periodization and computational evaluation of established periodizations provide some support for a break between an early and later Classical Arabic around the 1000s CE, which is somewhat earlier than typically believed, and suggests an additional previously unexplored division in the late 15th century CE. 

The processed corpus and the code associated with this work are 
available to the research community. We hope that future work will illuminate other aspects of the history of the Arabic language, as well as utilize the methods proposed in this work for studying other languages.

\newpage 

\bibliography{lrev-lt4dh}
\bibliographystyle{spmpsci}

\appendix

\section{Mathematical Notation}
\begin{tabular}{ll}
\toprule
Symbol & Meaning \\ 
\midrule
$K$ / $M$ / $G$ & $10^3$ / $10^6$ / $10^9$ \\
$\mathcal{T}$ & Collection of texts \\
$\mathcal{P}$ & List of texts organized in time periods \\
$P_i$ & Texts belonging to time period $i$ \\ 
$W_1$, $W_2$ & Word embedding matrices \\ 
$Q$, $U$, $V$, $\Sigma$ & Matrices \\   
$I$ & Identity matrix \\ 
$Q^T$ & Transpose of $Q$ \\ 
$||W||$ & Matrix norm \\
$||W||_F$ & Frobenius norm \\
\bottomrule
\end{tabular}

\section{Arabic examples} \label{sec:examples}
\paragraph{Boiler Plate Matches}
The following examples illustrate the boilerplate passages identified in the first step of the text reuse algorithm (Section~\ref{sec:reuse}).  The first is a chain of transmission that occurs 2,747 times in the corpus. The second is a part of a \gls{hadith} that occurs 221 times.  
\begin{lstlisting}
Hdvny mHmd bn Emrw qAl vnA >bw EASm qAl vnA EysY wHdvny AlHArv qAl vnA AlHsn qAl vnA wrqA' jmyEA En Abn >by njyH En mjAhd
\end{lstlisting}

\textit{Muhammad bin Amer told me, saying, ``Abu Asim told me that Isa told him that Al-Harith said, `Al-Hasan spoke to us, saying ``Warqa told all of us, based on what the son of Abi Najih said, quoting Mujahid"'"} 

\begin{lstlisting}
mn sy}At >EmAlnA mn yhdh Allh flA mDl lh wmn yDll flA hAdy lh w>$hd >n lA <lh <lA Allh wHdh lA $ryk lh w>$hd >n mHmdA Ebdh wrswlh
\end{lstlisting}
\textit{[I seek refuge in god] from the evil of our deeds; he who God guides rightly cannot go astray; he who goes astray cannot be led aright; I witness that there is no God but God alone with no partner, and that Muhammad is His servant and His Prophet} 

\paragraph{Near-matches}
The next example is of a near-match with several differences. The first is the original, from \texttt{0292Yacqubi.TarikhYacqubi.Shia003468Vols-ara1}:
\begin{lstlisting}
lys Tlby llElm TmEA fy blwg qASyth wAlAsjylA' ElY gAyth wlkn Altms $y}A lA ysE jhlh wlA yHsn bAlEAql.
\end{lstlisting}
\textit{My search for knowledge is not greed to reach its utmost, or to seize its aim. Rather, I seek something of which ignorance is widespread, and which the wise dare not contradict.}\\

As identified in a later text (\texttt{1371MuhsinCamili.AcyanShica.Shia003636Vols}), differences between forward slashes:
\begin{lstlisting}
lys Tlby llElm TmEA fy blwg qASyth wAlAsjylA' ElY /nhAyth/ wlkn /mErfp ma/ lA ysE jhlh wlA yHsn bAlEAql.
\end{lstlisting}
\textit{My search for knowledge is not greed to reach its end, or to seize its aim. Rather, knowledge of that of which ignorance is widespread and which the wise dare not contradict.}

\newpage
\setglossarystyle{long3col-booktabs}
\setlength{\glsdescwidth}{0.5\linewidth}
\setlength{\glspagelistwidth}{0.2\linewidth}
\printglossaries

\end{document}